%% file: cvpr2026_main.tex
\documentclass[10pt,twocolumn,letterpaper]{article}

\usepackage{cvpr}              

\input{preamble}

\definecolor{cvprblue}{rgb}{0.21,0.49,0.74}
\usepackage[pagebackref,breaklinks,colorlinks,allcolors=cvprblue]{hyperref}

\def\paperID{31825} 
\def\confName{CVPR}
\def\confYear{2026}

\title{Memory-Efficient Transfer Learning with Fading Side Networks via Masked Dual Path Distillation}

\author{
\textbf{Yutong Zhang}\textsuperscript{\rm 1,\rm 2}, \textbf{Jiaxin Chen}\textsuperscript{\rm 1,\rm 2}\thanks{Corresponding Author.}\;, \textbf{Honglin Chen}\textsuperscript{\rm 1,\rm 2}, \textbf{Kaiqi Zheng}\textsuperscript{\rm 1,\rm 2}, \textbf{Shengcai Liao}\textsuperscript{\rm 3},\\ \textbf{Hanwen Zhong}\textsuperscript{\rm 1,\rm 2}, \textbf{Weixin Li}\textsuperscript{\rm1,\rm 2}, \textbf{Yunhong Wang}\textsuperscript{\rm 1,\rm 2}\\
\textsuperscript{\rm 1}State Key Laboratory of Virtual Reality Technology and Systems, Beihang University, China\\
\textsuperscript{\rm 2}School of Computer Science and Engineering, Beihang University, China\\
\textsuperscript{\rm 3}College of Information Technology, United Arab Emirates University, United Arab Emirates\\
{\tt\small \{ytzhang\_mq, jiaxinchen, weixinli, yhwang\}@buaa.edu.cn, scliao@ieee.org}
}

\begin{document}
\maketitle
\input{main_sec/0_abstract}    
\input{main_sec/1_introduction}
\input{main_sec/2_related_work}
\input{main_sec/3_methodology}
\input{main_sec/4_experimental}

\input{main_sec/5_conclusion}

\clearpage
\input{main_sec/6_acknowledgment}
{
    \small
    \bibliographystyle{ieeenat_fullname}
    \bibliography{main}
}

\input{main_sec/X_suppl}

\end{document}

%% file: preamble.tex
\usepackage{amsfonts}
\usepackage{amsthm}
\usepackage{bm}
\usepackage{multirow}
\usepackage{makecell}
\usepackage{booktabs}
\usepackage{amsmath}
\usepackage{amssymb}
\usepackage{pifont}
\usepackage{algorithm}
\usepackage{algorithmic}
\usepackage[accsupp]{axessibility}



%% file: main_sec/0_abstract.tex
\begin{abstract}
Memory-efficient transfer learning (METL) approaches have recently achieved promising performance in adapting pre-trained models to downstream tasks. They avoid applying gradient backpropagation in large backbones, thus significantly reducing the number of trainable parameters and high memory consumption during fine-tuning. However, since they typically employ a lightweight and learnable side network, these methods inevitably introduce additional memory and time overhead during inference, which contradicts the ultimate goal of efficient transfer learning. To address the above issue, we propose a novel approach dubbed Masked Dual Path Distillation (MDPD) to accelerate inference while retaining parameter and memory efficiency in fine-tuning with fading side networks. Specifically, MDPD develops a framework that enhances the performance by mutually distilling the frozen backbones and learnable side networks in fine-tuning, and discard the side network during inference without sacrificing accuracy. Moreover, we design a novel feature-based knowledge distillation method for the encoder structure with multiple layers. Extensive experiments on distinct backbones across vision/language-only and vision-and-language tasks demonstrate that our method not only accelerates inference by at least 25.2\% while keeping parameter and memory consumption comparable, but also remarkably promotes the accuracy compared to SOTA approaches. {The source code is available at {\small \url{https://github.com/Zhang-VKk/MDPD}}.}
\end{abstract}

%% file: main_sec/1_introduction.tex
\vspace{-0.6cm}
\section{Introduction}
\label{introduction}
Recently, large-scale pre-trained foundation models have demonstrated remarkable representation and generalization capabilities across diverse domains, including computer vision \cite{fang2023eva,he2022masked}, natural language processing \cite{radford2019language,touvron2023llama,zhuang2024towards}, and multi-modal tasks \cite{li2023blip,sun2023generative,diao2024unveiling,liang2024simple,zhang2026mingfang}. To harness strengths of these well-established models and enhance their performance towards downstream tasks, one of the most commonly used strategies is fully fine-tuning parameters of these models based on downstream datasets. However, as the number of parameters continues to vastly grow, this straightforward method becomes prohibitively computationally expensive and is prone to over-fitting, especially considering that the training data for downstream tasks is typically limited in scale. 

To deal with the above issues, parameter-efficient transfer learning (PETL) approaches \cite{houlsby2019param,hu2022lora,zhong2026hanwen} have been proposed by tuning a small number of parameters or inserting additional lightweight modules. Existing PETL methods can be roughly divided into the following three categories: 1) {Partially Tuning} \cite{guo2020parameter,sung2021training} updates a small subset of task-specific parameters while freezing the majority of the original backbone. 2) {Prompt Tuning} \cite{jia2022visual,zhou2022learning,li2021prefix,li2024multimodal} introduces a fixed number of learnable prompt vectors, which are subsequently optimized while all original parameters remain frozen. 3) {Adapter Tuning} \cite{chen2022adaptformer,chen2022vision,liao2024uni,yin20231,yin2024parameter,yin20255} incorporates additional bottleneck-shaped modules into the backbone, updating them while the backbone stays frozen. These methods significantly reduce the number of trainable parameters while achieving performance comparable to full fine-tuning. However, since the gradients are computed throughout almost the entire backbone, they still incur substantial memory consumption during training, significantly restricting their practical applicability.

Memory-efficient transfer learning (METL) \cite{zhang2020side,sung2022lst,lin2023hierarchical,mercea2024time,liu2024tuning} has recently emerged as a promising approach to achieve both parameters and memory efficiency during training. Typically, METL involves constructing a lightweight and learnable parallel side network alongside the backbone network \cite{diao2024unipt,diao2024sherl}, with static intermediate feature pairs integrated between the two at each layer. This fine-tuning paradigm reduces memory overhead by updating only the \textit{small-scale} side network while freezing the \textit{large-scale} backbone, and computing gradients exclusively for the side network during back-propagation. However, the additional forward propagation introduced by the side network result in increased inference time and memory costs, which conflicts with the ultimate goal of achieving both efficient fine-tuning and inference. 

To address the aforementioned limitations, we propose a novel approach dubbed Masked Dual Path Distillation (MDPD) for memory-efficient transfer learning based on the side network framework. Without compromising inference speed, MDPD distills intermediate features from the backbone and the final output of the side network separately, while efficiently guiding and updating the backbone through the lightweight side network, thereby reducing training memory overhead. Specifically, during training, feature- and logits-based knowledge distillation methods are applied to the intermediate features of the backbone and the final logits output of the side network. In a dual-path framework, alternating optimization of the backbone and the side network minimizes their discrepancy in feature distributions. During inference, only the optimized backbone is utilized, eliminating additional inference time while maintaining the performance, thus achieving efficiency in both training memory and inference speed. Furthermore, previous feature distillation strategies often require the student model to directly mimic the intermediate features of the teacher model. Nevertheless, the feature distributions of these two models typically differ, particularly in the attentive areas of the input, making direct imitation challenging. Additionally, existing methods usually only distill a subset of the layers or even just the final layer of the teacher or student network. For backbones with multiple encoding layers such as ViT and BERT, where deep and shallow layers focus on different tokens, it is equally important to perform distillation in the deep and shallow layers. Therefore, We also propose a novel feature-based distillation strategy by applying direct imitation in the shallow layer of the backbone, while employing mask generation in the deep layers, ensuring more effective and balanced distillation.


The main contributions of this work lie in three-fold:

(1) We propose Masked Dual Path Distillation (MDPD), a novel approach designed to accelerate current METL frameworks, which introduces the side networks to guide backbone optimization with mutual distillation during training, while discarding them during inference to enhance efficiency without compromising accuracy.

(2) We develop a Hierarchical Feature-based Distillation method by employing tailored  distillation strategies for distinct layers to mitigate the discrepancies between the backbone and side networks. 

(3) We extensively evaluate our method across multiple visual-language and vision/language-only tasks based on distinct backbones, demonstrating that our approach remarkably reduces the inference time by at least 22.5\%, while maintaining accuracy as well as memory and parameter efficiency.

%% file: main_sec/2_related_work.tex
\section{Related Work}
\label{related_work}

\subsection{Parameter-Efficient Transfer Learning}
Pre-trained foundation models have achieved remarkable success across various domains~\cite{devllin2019bert, liu2019roberta, alexey2021vit, radford2021learning, girdhar2023image,wu2026zimeng}. However, their application to downstream tasks encounters challenges in computational efficiency and adaptability, sparking significant interest in parameter-efficient transfer learning (PETL) techniques. Existing PETL methods roughly fall into the following categories. 1) \emph{Selective tuning} methods update only a subset of parameters, such as bias \cite{cai2020tinytl, zaken2022bitfit}, weights \cite{zhao2020masking, touvron2022three} and normalization layers \cite{kim2021howto}, while freezing the rest. 2) \emph{Additive methods} introduce additional parameters and focus on training them instead of backbones, including Adapter-based ones \cite{houlsby2019param, zhu2021counter, he2022sparse, he2022towards, he2023sensitivity,chen2026zhenghao} and Prompt tuning based ones \cite{hambardzumyan2021warp, lester2021power, li2021prefix, vu2021spot, qin2022exploring}. (3) \emph{Re-parameterized} methods \cite{hu2022lora, he2023param, song2024increasing, wu2024mixture, wu2025unified,zhong2024hanwen} employ extra trainable parameters with low-rank structures to reduce computational cost.

\subsection{Memory-Efficient Transfer Learning}
Different from PETL that aims to reduce trainable parameters, memory-efficient transfer learning (METL) focuses on decreasing memory consumption during training. Typically, \cite{zhang2020side, sung2022lst, jiang2024res, diao2024unipt, diao2024sherl, zhang2026moismoe} accomplish memory-efficient tuning by introducing learnable side network parallel to the backbone \cite{zhang2020side, sung2022lst, jiang2024res, diao2024unipt, diao2024sherl}, which confine back-propagation to side network. It avoids storing gradients of large-scale backbones, thereby reducing memory overhead. 
Alternatively, \cite{liao2023make}, leverages reversible models \cite{gomez2017reversible, zhang2023lora-fa} by enabling activations to be recomputed during back-propagation rather than being stored during the forward pass. Some works \cite{malladi2023tune, phang2023hypertuning} attempt to avoid back-propagation.
Besides, \cite{gomez2017reversible,2016Training}
reconstruct discarded activations from backward layers or perform gradient checkpoint operation without storing all intermediate activations, and \cite{gomez2017reversible}. Despite memory efficient, these methods significantly limit the model capacity and lag far behind the performance of full fine-tuning.

\subsection{Knowledge Distillation}
Knowledge distillation (KD) \cite{hinton2015distilling,pan2024disentangled} trains a smaller student model to mimic the larger teacher model, which can be categorized into logits-based, feature-based and relation-based methods. 1) \emph{Logits-based} methods \cite{meng2019conditional, wu2022tinyvit} focus on mimicing the final output of the teacher model.
2) \emph{Feature-based} methods \cite{romero2015fitnets, chen2021cross, bai2023masked} leverage the intermediate representations of  teacher networks to guide the learning process of student networks.
3) \emph{Relation-based} methods \cite{yim2017gift, chen2021learning, chen2024joint} explore relationships between different layers of teacher models as knowledge, which is utilized to facilitate learning of student models.

%% file: main_sec/3_methodology.tex
\begin{figure*}[!t]
\centering
\includegraphics[width=0.98\textwidth]{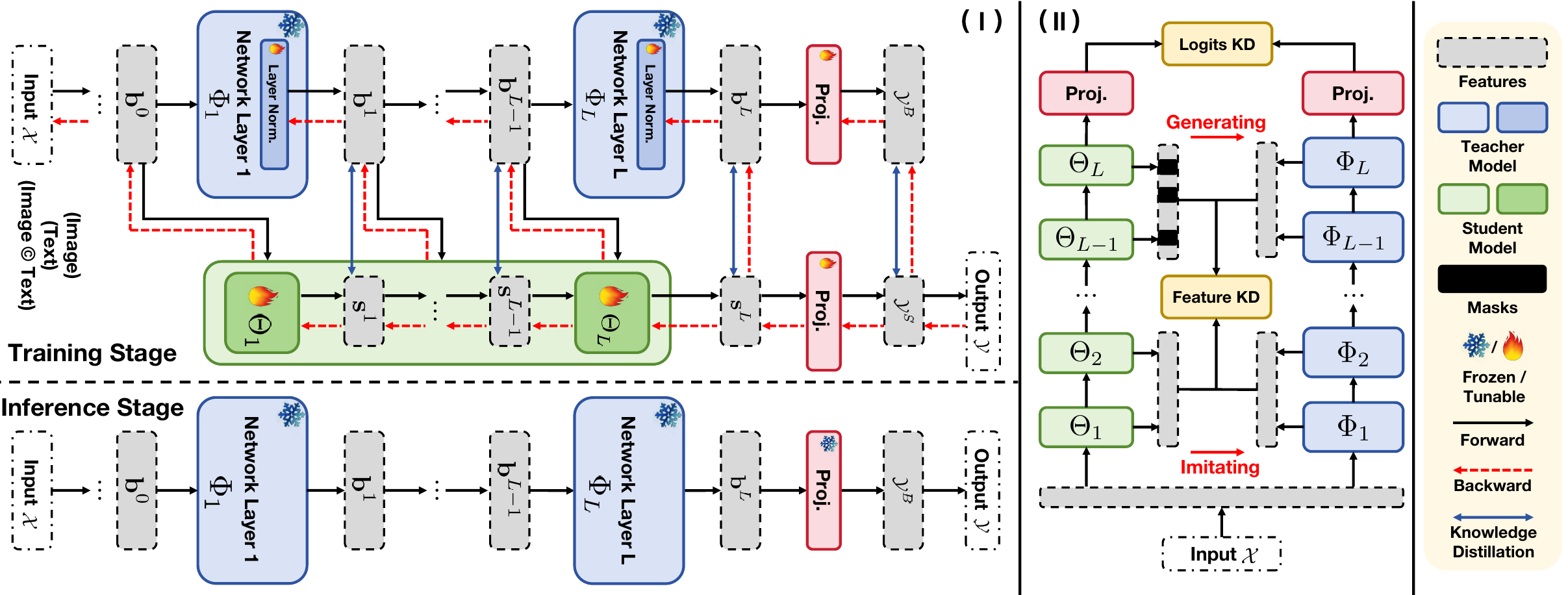} 
\caption{Illustration on our method. Part (I) shows the pipeline of Masked Dual Path Distillation. During training, the backbone and side networks mutually act as teacher and student for effective distillation, where side networks are abandoned after training for inference efficiency. Part (II) displays Hierarchical Feature-based Distillation that adopt distinct strategies across different layers. It particularly selects a set of masked features, making them learnable with guidance of backbones in deep layers. 
}
\label{fig-framework}
\vspace{-0.3cm}
\end{figure*}

\section{Methodology}
\subsection{Framework Overview}
\label{overview}
Generally, previous METL methods significantly reduce training memory overhead by constructing a lightweight, trainable side network to avoid gradients directly passing through a large backbone. Similar to the structure of backbone, side network is usually composed of several intermediate layers and initialized by pre-trained weights. However, the corresponding layers of two networks cannot be fully forward propagated parallelly, as the input of the side network branch is guided by the output of the intermediate layer or final layer of backbone. Therefore, during the forward pass of the model, the side branch inevitably introduces additional computational overhead in both training and inference, ultimately reducing the overall efficiency. To address this issue, we propose a novel method dubbed Masked Dual Path Distillation (MDPD) as shown in Fig.~\ref{fig-framework}.

Specifically, given a single modality input $\mathcal{X}$, a pre-trained backbone $\Phi_{1:L}$ containing $L$ layers is used to extract $L$ token sequences $\bm{b}^1,\cdots,\bm{b}^L\in\mathbb{R}^{N\times D_B}$, each consisting of $N$ tokens with a hidden dimensionality of $D_B$. After obtaining final output $\bm{b}^L$ of backbone, it is fused with each intermediate sequences $\bm{b}^l,l\in 1,2,\cdots,L-1$ and regarded as corresponding layer input of side network, which also contains $L$ layers $\Theta_{1:L}$, to extract side token sequences $\bm{s}^1,\cdots,\bm{s}^L\in \mathbb{R}^{N\times D_S}$, usually consisting of $N$ $D_S$-dimensional ($D_S{<D_B}$) tokens. Subsequently, in order to utilize powerful model representation ability of pre-trained network, backbone and side network is regarded as teacher $\mathcal{T}_{fea}$ and student $\mathcal{S}_{fea}$ model respectively, and intermediate features of corresponding layers $\bm{b}^l$ and $\bm{s}^l$ are distilled to enhance feature extraction ability of side branch. In particular, when performing feature-based distillation on an encoder structure which containing multiple layers, we find that there are significant differences in attention pattern of teacher-student network at different layers. For shallow layers of the network, attention map of students and teachers is diagonal, showing a \textit{self-attention} pattern; while for deep layers, attention patterns gradually diverge, and begin to focus on different sparse key tokens respectively. This difference makes it difficult for students to directly imitate distribution of teacher's final features. Therefore, in response to above findings, we fine-grain knowledge distillation strategies for different layers. Specifically, for shallow layers with smaller differences, students are forced to directly mimic features of corresponding layers of teacher and learn how to focus on the tokens themselves; for deep layers with larger differences and stronger semantic information, students are forced to generate features of teacher rather than directly imitate them. In addition, since only final layer features of side network are used as the final output of the model during the training stage, and backbone is not considered (\emph{i.e.}, the backbone is not followed by a task-specific head), we manually design a new output layer, while regard backbone and side network as a student $\mathcal{S}_{log}$ and teacher $\mathcal{T}_{log}$ model respectively for logits-based distillation, which is performed on final outputs of two networks $\mathbf{b}_L$ and $\mathbf{s}_L$ and transfer knowledge of side network to backbone to optimize latter's model output performance. {At training stage, most of parameters in backbone remain frozen to prevent a large amount of gradients from flowing through, and only scaling and shifting coefficients of Layer Normalization in each block and parameters of final output layer are updated, while all parameters in side network are learnable as well.} In inference stage, backbone equipped with a task-specific head is directly used to perform inference operations, while side branch is discarded directly, thereby achieving parameter- and memory-efficient training and time-efficient inference, and gain better performance.

\subsection{Dual Path Knowledge Distillation}
\label{dual_path}

In order to enable backbone to have the ability of independent inferring, we propose a Dual Path Knowledge Distillation (DPKD) strategy, which aims to distill the downstream task-specific knowledge learned by the side network into backbone, so as to be efficiently guided and updated. In order to make full use of prior knowledge of pre-trained backbone and enable side network to obtain a stronger model representation capacity, based on knowledge distillation, we first adopt feature-based distillation to calculate distillation loss for corresponding intermediate layer features of backbone and side network. Specifically, backbone $B$ and side network $S$ are regarded as teacher $\mathcal{T}_{fea}$ and student $\mathcal{S}_{fea}$, respectively. After obtaining $l$-th layer features of backbone $\bm{b}^l\in\mathbb{R}^{N\times D_B}$ and side branch $\bm{s}^l\in\mathbb{R}^{N\times D_S} (D_S<D_B)$, hidden dimensions need to be aligned firstly in order to perform distillation operation. However, if a $1 \times 1$ convolutional layer is used to implement alignment operation, approximately $(1+D_B)\times D_S$ parameters will be introduced, which violates the goal of parameter-efficiency. Therefore, we consider using two low-rank matrices $\bm{M}_{down}\in \mathbb{R}^{D_S\times d}$ and $\bm{M}_{up}\in \mathbb{R}^{d\times D_B}$ in the form of a bottleneck, first reducing $D_S$ to a lower dimension $d (d\ll \min(D_B,D_S))$, and then increasing it to $D_B$:
\begin{equation}
    \phi(\bm{s}^l)=\bm{s}^l\bm{M}_{down}\bm{M}_{up} \Rightarrow \bm{b}^l,
\end{equation}
where $l\in 1,2,\cdots,12$ represents the layer index, $\phi(\cdot)$ is the bottleneck module used to reshape the dimension of $\bm{s}^l$ to the same as $\bm{b}^l$. This will only introduce about $(1+D_S+D_B)\times d+D_B$ parameters, but still be able to obtain comparable or even better experimental results than a single convolutional layer, while ensuring parameter- and memory-efficiency. Then, we calculate the distillation loss $\mathcal{L}_{\text{fea}}$ for the intermediate features of each layer:
\begin{equation}\label{feature_loss}
    \mathcal{L}_{\text{fea}}^l=\sum_{i=1}^{N}\sum_{j=1}^{D_B}(\bm{b}^l_{i,j}-\phi(\bm{s}^l)_{i,j})^2,
\end{equation}
where $N$,$D_B$ represent the number of tokens and hidden dimensions of teacher's feature, respectively. After feature-based distillation, side branch obtains stronger representation capability under the guidance of pre-trained backbone, thus can mine richer and more important features.

In addition, as the output of side network is final output of the model without the contribution of backbone, so as to ensure that only the backbone is used for inferring, a strong task-related mapping layer is required to obtain final logits. We first use global average pooling (GAP) on final feature maps of backbone and side branch $\bm{b}^L$ and $\bm{s}^L$, then adopt linear mapping layers $\bm{W}_B$ and $\bm{W}_S$ to get output logits of two networks $\mathcal{Y}^B\in\mathbb{R}^{D_B'}$ and $\mathcal{Y}^S\in\mathbb{R}^{D_S'}$, respectively. Successively, side network and backbone are regarded as teacher $\mathcal{T}_{log}$ and student $\mathcal{S}_{log}$ respectively, and we adopt logits-based distillation strategy to calculate loss $\mathcal{L}_{\text{log}}$ for $\mathcal{Y}^B$ and $\mathcal{Y}^S$: 
\begin{equation}\label{logits_loss}
    \mathcal{L}_{\text{log}}=\sum_{i=1}^{D_S'}(\mathcal{Y}^S_{i}-\phi(\mathcal{Y}^B)_{i})^2,
\end{equation}
After logits-based distillation, the output layer of the backbone can independently obtain stronger mapping ability in the inference stage under the guidance of the side branch, thereby showing better model performance.

{In the above process, the backbone $B$ and the side network $S$ take turns as teacher or student models. Through the Dual Path Knowledge Distillation method, they teach each other the knowledge of their own strong parts and learn from each other the knowledge of their own weak parts. The two guide each other and achieve alternating optimization.} Therefore, in the inference stage, only the backbone is used for inferring and the side network is discarded. While avoiding introducing additional inference time, the model can still maintain a better performance, thereby achieving a balance between performance, parameters, memory, and time. 

\subsection{Hierarchical Feature-based Distillation}
\label{feature_distill}

As teacher and student networks typically exhibit similar attention patterns in shallow layers, we guide student to align with teacher's distribution. Specifically, we adopt the following loss in shallow layers:
\begin{equation}
    \mathcal{L}_{\text{sha}}^l=\sum_{i=1}^{N}\sum_{j=1}^{D_B}(\bm{b}^l_{i,j}-\phi(\bm{s}^l)_{i,j})^2.
\end{equation}

In deep layers, features of teacher and student networks often have significantly varying attention patterns, making direct distillation infeasible. Therefore, we try to force the student to fit the distribution of the teacher's features through generation. Generally, deep features contain rich semantic information. At the same time, the tokens of such features also contain information about adjacent tokens. Therefore, we use the student's masked features to generate the teacher's features, which only use partial tokens to represent the complete original attention map. This can not only retain some of the original student feature information, but also obtain a better feature representation through generation. Similarly, we first use a bottleneck module to align the hidden dimensions of the student's and teacher's features. {Then, we set a set of shared random mask values $\bm{m}\in\mathbb{R}^{N\times 1}$ to specify the coordinates of the student's original feature tokens replaced by the learnable mask tokens:}
\begin{equation}
\label{eq:mask}
    \bm{m}_i=
    \begin{cases}
        1,\enspace {\rm{if}}~~~r_i<\lambda, \\
        0,\enspace \rm{otherwise},
    \end{cases}
\end{equation}
where $r_i\in [0,1]$ is a random number deciding whether the $i$-th token is masked. {$\lambda$ is the ratio controlling the proportion of masking, which is unitarily applied to all deep layers of side network.} Based on $\bm{m}_i$, we update features of student network as below:
\begin{equation}
    \bar{\bm{s}}_i=
    \begin{cases}
        \bm{f}_{mask},\enspace {\rm{if}}~~~\bm{m}_i, \\
        \bm{f}_{ori},\enspace \rm{otherwise},
    \end{cases}
\end{equation}
where $\bm{f}_{ori}$ is the the original features and  $\bm{f}_{mask}$ is the learnable masked feature. Then, we utilize $\bar{\bm{s}}$ to generate the teacher's features with a generation block $\mathcal{G}(\cdot)$, which is a projector with two 3$\times$3 convolutional layers and one ReLU activation layer. Finally, the loss of feature-based distillation in deep layers $\mathcal{L}_{\text{deep}}$ is expressed as:
\begin{equation}
\label{eq:generation_block}
\mathcal{L}_{\text{deep}}^l=\sum_{i=1}^{N}\sum_{j=1}^{D_B}\bm{m}_i\cdot (\bm{b}^l_{i,j}-\mathcal{G}(\bar{\bm{s}^l})_{i,j})^2.
\end{equation}

%% file: main_sec/4_experimental.tex
\section{Experimental Results and Analysis}
\label{experiment}

\subsection{Experimental Setup}

\textbf{Dataset.}
We primarily evaluate performance of our method on various \emph{vision-language (VL) tasks}, including:
(1) Image-Text Retrieval ($\textbf{ITR}$) on Flickr30K \cite{young2014image} and MSCOCO \cite{lin2014microsoft}; (2) Video-Text Retrieval ($\textbf{VTR}$) on MSVD \cite{chen2011collecting} and MSR-VTT \cite{xu2016msr}; (3) Question Answering ($\textbf{VQA\&GQA}$) on VQAv2 \cite{goyal2017making} and GQA \cite{hudson2019gqa}; and (4) Visual Grounding ($\textbf{VG}$) on RefCOCO, RefCOCO+ \cite{yu2016modeling} and RefCOCOg \cite{mao2016generation}. Additionally, we evaluate our approach on \emph{vision-only and language-only tasks} on \textbf{VTAB-1K} \cite{zhai2019large} and \textbf{GLUE} benchmark \cite{wang2018glue}, respectively. 

\noindent\textbf{Evaluation Metrics.}
We report Recall@1 and Rsum on ITR and VTR tasks, overall Accuracy on QA tasks, and mean Average Precision on VG tasks. For GLUE benchmark, we present Accuracy Metric, F1 Score, Matthew's Correlation, Pearson-Spearman Correlation as the evaluation metrics for various datasets respectively. Besides, we state Top-1 Accuracy on 19 datasets in VTAB-1K. More details are in \textit{Supplementary Material}. 
Besides, we report GPU memory usage (G) during training, number of trainable parameters (M), and inference latency (ms per run), FPS (frames per second) or QPS (queries per second) as efficiency metrics.

\noindent\textbf{Implementation Details.}
To ensure fair comparisons, we follow \cite{diao2024unipt,diao2024sherl} by aligning most settings with the original pre-trained models, including the optimizer, warm-up scheduler, batch size, training epochs, \emph{etc}. More details are provided in \textit{Supplementary Material}.


\begin{table*}[t]
    \centering
    \caption{Comparison results with PETL \textbf{(Middle)} and METL \textbf{(Bottom)} approaches across distinct network architectures. Best results are in \textbf{bold}, and second best results are \underline{underlined}.}
    \vspace{-0.1cm}
    \small
    \setlength{\tabcolsep}{1.9pt}
    \scalebox{0.95}{
    \begin{tabular}{lcccccc|cccccc|cccccc}
    \toprule
    
    \multirow{2}{*}{Method} & Params. & Mem. & QPS & \multicolumn{3}{c|}{BERT+BUTD} & Params. & Mem. & QPS & \multicolumn{3}{c|}{ResNeXt+BiGRU} & Params. & Mem. & QPS & \multicolumn{3}{c}{ViT+Text Transf.} \\
        \cmidrule[0.4pt]{5-7} \cmidrule[0.4pt]{11-13} \cmidrule[0.4pt]{17-19}
        & (M) $\!\downarrow$ & (G) $\!\downarrow$ & $\!\uparrow$ & I-T $\!\uparrow$ & T-I $\!\uparrow$ & Rsum $\!\uparrow\ $ & (M) $\!\downarrow$ & (G) $\!\downarrow$ & $\!\uparrow$ & I-T $\!\uparrow$ & T-I $\!\uparrow$ & Rsum $\!\uparrow\ $ & (M) $\!\downarrow$ & (G) $\!\downarrow$ & $\!\uparrow$ & T-V $\!\uparrow$ & V-T $\!\uparrow$ & Rsum $\!\uparrow$ \\
    
    \midrule

    \textcolor{gray}{Fully-FT} & \textcolor{gray}{109.5} & \textcolor{gray}{9.9} & \textcolor{gray}{24.14} & \textcolor{gray}{79.7}  & \textcolor{gray}{62.1} & \textcolor{gray}{513.5}  & \textcolor{gray}{90.9} & \textcolor{gray}{174.4} & \textcolor{gray}{30.80} & \textcolor{gray}{85.6} & \textcolor{gray}{70.2} & \textcolor{gray}{539.0}  & \textcolor{gray}{151.3} & \textcolor{gray}{48.8} & \textcolor{gray}{44.07} & \textcolor{gray}{42.8} & \textcolor{gray}{42.1} & \textcolor{gray}{389.2} \\
        \textcolor{gray}{Partially} & \textcolor{gray}{0.8} & \textcolor{gray}{1.0} & \textcolor{gray}{24.19} & \textcolor{gray}{74.8} & \textcolor{gray}{57.3} & \textcolor{gray}{485.5}  & \textcolor{gray}{2.1} & \textcolor{gray}{14.9} & \textcolor{gray}{31.26} & \textcolor{gray}{75.2} & \textcolor{gray}{58.2} & \textcolor{gray}{505.8} & \textcolor{gray}{0.7} & \textcolor{gray}{7.6} & \textcolor{gray}{44.48} & \textcolor{gray}{36.4} & \textcolor{gray}{37.0} & \textcolor{gray}{353.9} \\
        \midrule[0.6pt]
        Adapter \cite{houlsby2019param} & 2.6 & 8.8 & 22.44 & 79.1  & \underline{60.5} & \underline{511.3}  & 3.5 & 176.8 & 28.65 & 66.8 & 62.9 & 493.3  & 5.2 & 41.2 & 39.17 & 38.3 & 39.6 & 364.3 \\
        LoRA \cite{hu2022lora} & \underline{1.1} & 8.8 & 23.37 & 78.8 & 59.6 & 508.2  & -- & -- & -- & -- & -- & --  & 1.3 & 40.8 & 41.55 & 38.8 & 39.9 & 366.8 \\
        BitFit \cite{zaken2022bitfit}  & 0.9 & 8.6 & \underline{23.70} & 77.3 & 57.8 & 503.9  & \underline{2.2} & 168.0 & \underline{30.53} & 83.4 & 67.4 & 530.6 & \textbf{0.1} & 42.0 & \underline{43.12} & 38.1 & 40.6 & \underline{370.8} \\
        Prompt \cite{li2021prefix} & 10.7 & 9.4 & 17.64 & 78.7 & 59.0 & 508.5  & -- & -- & -- & -- & -- & --  & \underline{0.2} & 42.8 & 33.74 & 36.8 & 37.5 & 358.8 \\
        SSF \cite{lian2022scaling} & \textbf{0.2} & 8.4 & 22.23 & 80.0 & 60.4 & 512.8  & \textbf{0.1} & 163.2 & 27.63 & 83.7 & 66.8 & 528.5  & 0.5 & 39.2 & 37.91 & \textbf{40.2} & \textbf{41.8} & \textbf{376.6} \\
        FacT \cite{jie2023fact} & 0.6 & 8.7 & -- & 79.2 & 59.3 & 508.8  & -- & -- & -- & -- & -- & --  & 0.8 & 40.8 & -- & 38.7 & 39.8 & 367.2 \\
        AdaLoRA \cite{zhang2023adaptive} & 1.0 & 8.8 & 23.03 & 79.8 & 60.1 & 510.3  & -- & -- & -- & -- & -- & --  & 1.2 & 42.0 & 40.06 & \underline{39.2} & 39.6 & 368.5 \\
        \midrule[0.6pt]
        LST \cite{sung2022lst} & 7.5 & 4.6 & 10.73 & 77.9 & 57.3 & 501.9 & 2.3 & \textbf{15.0} & 13.86 & 82.3 & 66.1 & 526.7  & 11.2 & 32.0 & 19.85 & 37.0 & 37.8 & 356.7 \\
        UniPT \cite{liao2024uni} & 5.9 & \textbf{3.1} & 12.75 & \underline{80.2} & 59.8 & 510.5 & 6.4 & \textbf{15.0} & 17.83 & \underline{84.0} & \underline{67.7} & \underline{532.1}  & 9.6 & \textbf{13.6} & 27.03 & 38.9 & 39.3 & 361.3 \\
        
        \textbf{Ours} & 5.9  & \underline{3.4} & \textbf{24.05} & \textbf{81.9} & \textbf{61.1} & \textbf{516.2} & 6.5 & \underline{15.7} & \textbf{30.62} & \textbf{85.3} & \textbf{69.8} & \textbf{539.1} & 9.7 & \underline{14.0} & \textbf{43.52} & \textbf{40.2} & \underline{40.7} & 365.8 \\
    
    \bottomrule
    \end{tabular}
    }
    \label{tab1}
    \vspace{-0.2cm}
\end{table*}

\begin{table*}[!t]
    \centering
    \caption{Comparison results with PETL \textbf{(Top)} and METL \textbf{(Middle, Bottom)} methods on GLUE. Ours${}^\heartsuit$ and Ours${}^\spadesuit$ indicate our method implemented based on UniPT and SHERL, respectively. Best results are in \textbf{bold}, and second best results are \underline{underlined}.}
    \vspace{-0.3cm}
    \begin{center}
    \resizebox{\textwidth}{!}{
    \begin{tabular}{lccccccccccccc}
         \toprule[1.2pt]
         \multirow{2}{*}{\centering Method} & Params. & \multicolumn{2}{c}{Memory (G) $\downarrow$} & Time & \multirow{2}{*}{CoLA} & \multirow{2}{*}{SST-2} & \multirow{2}{*}{MRPC} & \multirow{2}{*}{QQP} & \multirow{2}{*}{MNLI} & \multirow{2}{*}{QNLI} & \multirow{2}{*}{RTE} & \multirow{2}{*}{STS-B} & \multirow{2}{*}{Mean} \\
         & (\%) $\downarrow$ & Train & Test & (ms) $\downarrow$ & & & & & & & & & \\
         \midrule[0.6pt]
         \textcolor{gray}{Fully-FT} & \textcolor{gray}{100} & \textcolor{gray}{17.6} & \textcolor{gray}{0.86} & \textcolor{gray}{232.5} & \textcolor{gray}{62.8} & \textcolor{gray}{93.9} & \textcolor{gray}{91.9} & \textcolor{gray}{89.9} & \textcolor{gray}{86.2} & \textcolor{gray}{92.5} & \textcolor{gray}{74.1} & \textcolor{gray}{90.3} & \textcolor{gray}{85.2} \\
         \hline
         Adapter \cite{houlsby2019param} & 1.63 & 13.0 & 0.87 & 247.3 & 64.4 & 94.2 & 88.9 & 88.9 & 86.4 & 93.1 & 75.1 & 91.1 & 85.3 \\
         LoRA \cite{hu2022lora} & 1.71 & 12.6 & 0.86 & 238.1 & 63.3 & 94.3 & 90.1 & 89.0 & 86.3 & 93.2 & 75.5 & 90.9 & 85.3 \\
         BitFit \cite{zaken2022bitfit} & 0.13 & 10.7 & 0.86 & 236.9 & 61.8 & 94.3 & 91.0 & 88.7 & 85.6 & 93.1 & 67.6 & 90.8 & 84.1 \\
         Prompt \cite{li2021prefix} & 0.03 & 22.2 & 0.87 & 291.4 & 0 & 90.3 & 74.6 & 88.5 & 82.5 & 92.5 & 59.5 & 90.1 & 72.2 \\
         \midrule[0.6pt]
         LST \cite{sung2022lst} & 1.74 & 5.5 & 0.88 & 462.7 & 58.1 & 94.1 & 90.4 & 88.8 & \underline{85.6} & 93.3 & \underline{71.9} & 90.7 & 84.1 \\
         UniPT \cite{liao2024uni} & 1.36 & \textbf{2.9} & \textbf{0.86} & 423.6 & \underline{62.2} & \underline{94.2} & \underline{90.8} & 88.9 & 85.5 & 93.3 & 69.8 & 89.7 & 84.3 \\
         SHERL \cite{diao2024sherl} & \textbf{0.85} & \textbf{2.9} & \underline{0.87} & 350.4 & 61.1 & 93.7 & 89.4 & 88.8 & 85.3 & 93.3 & \underline{71.9} & \underline{90.9} & 84.3 \\
         \textbf{Ours${}^\heartsuit$} & 1.48 & \underline{3.3} & \textbf{0.86} & \underline{235.2} & \textbf{62.9} & \textbf{94.4} & \textbf{91.2} & \underline{89.0} & 85.2 & \underline{93.4} & 70.5 & 90.2 & \underline{84.6} \\
         \textbf{Ours${}^\spadesuit$} & \underline{1.02} & 3.4 & \underline{0.87} & \textbf{234.8} & 62.0 & 94.1 & 89.8 & \textbf{89.2} & \textbf{85.7} & \textbf{93.5} & \textbf{72.6} & \textbf{91.2} & \textbf{84.8} \\
         \midrule[0.6pt]
         LST~(T5-large) \cite{sung2022lst} & 1.23 & 12.2 & 2.88 & 832.9 & 65.3 & 95.7 & 91.6 & 89.7 & \underline{88.6} & 94.1 & 79.9 & \textbf{92.4} & 87.1 \\
         UniPT~(T5-large) \cite{liao2024uni} & 0.92 & 9.1 & \underline{2.82} & 795.1 & 65.7 & 95.8 & 92.0 & 89.7 & 88.2 & 94.2 & 79.6 & 92.0 & 87.2 \\
         SHERL~(T5-large) \cite{diao2024sherl} & \textbf{0.64} & \textbf{7.1} & \textbf{2.80} & 667.4 & 65.6 & 95.8 & \underline{92.9} & 89.6 & \underline{88.6} & 94.2 & \underline{80.8} & 92.1 & 87.5 \\
         \textbf{Ours${}^\heartsuit$(T5-large)} & 1.02 & 10.2 & \underline{2.82} & \underline{418.7} & \textbf{66.3} & \underline{96.3} & 92.8 & \underline{90.0} & 88.4 & \underline{94.5} & 79.9 & \underline{92.3} & \underline{87.6} \\
         \textbf{Ours${}^\spadesuit$(T5-large)} & \underline{0.81} & \underline{8.0} & \textbf{2.80} & \textbf{415.3} & \underline{66.2} & \textbf{96.5} & \textbf{93.5} & \textbf{90.2} & \textbf{88.9} & \textbf{94.7} & \textbf{81.5} & \textbf{92.4} & \textbf{88.0} \\
         \bottomrule[1.2pt]
    \end{tabular}
    }
    \end{center}
\vspace{-0.2cm}
    \label{tab_glue_results}
\end{table*}

\begin{table*}[!ht]
	\centering
     \caption{Comparison results with PETL \textbf{(Top)} and METL \textbf{(Bottom)} methods on VTAB-1K. Best results are in \textbf{bold}, and second best results are \underline{underlined}.}
     \vspace{-0.1cm}
        \small
	\setlength{\tabcolsep}{0.23pt}
	\begin{tabular}{
            p{1.8cm}<{}
            p{0.66cm}<{\centering} p{0.66cm}<{\centering} p{0.9cm}<{\centering}|
            p{0.66cm}<{\centering}p{0.66cm}<{\centering}p{0.66cm}<{\centering}p{0.66cm}<{\centering}p{0.66cm}<{\centering}p{0.66cm}<{\centering}p{0.66cm}<{\centering}|
            p{0.66cm}<{\centering}p{0.66cm}<{\centering}p{0.66cm}<{\centering}p{0.66cm}<{\centering}|
            p{0.66cm}<{\centering}p{0.66cm}<{\centering}p{0.66cm}<{\centering}p{0.66cm}<{\centering}p{0.66cm}<{\centering}p{0.66cm}<{\centering}p{0.66cm}<{\centering}p{0.66cm}<{\centering}| 
            p{0.66cm}<{\centering}
        }
	\toprule[1.5pt]
 
	\multicolumn{4}{c|}{}&\multicolumn{7}{c|}{\textbf{Natural}}&\multicolumn{4}{c|}{\textbf{Specialized}}&\multicolumn{8}{c|}{\textbf{Structured}}&\\
    
        {{\rotatebox{90}{\textbf{Method}}}}
        &{{\rotatebox {90}{Params.~(M) $\downarrow$}}}
        & {{\rotatebox {90}{Memory~(G) $\downarrow$}}}
        &{{\rotatebox {90}{FPS $\uparrow$}}}
	& {{\rotatebox {90}{Cifar100}}}
	& {{\rotatebox {90}{Caltech101}}}
	& {{\rotatebox {90}{DTD}}}
	& {{\rotatebox {90}{Flower102}}}
	& {{\rotatebox {90}{Pets}}}
	& {{\rotatebox {90}{SVHN}}}
	&{{\rotatebox {90}{Sun397}}}
	& {{\rotatebox {90}{Camelyon}}}
	& {{\rotatebox {90}{EuroSAT}}}
	& {{\rotatebox {90}{Resisc45}}}
	&{{\rotatebox {90}{Retinopathy}}}
	& {{\rotatebox {90}{Clevr-Count}}}
	& {{\rotatebox {90}{Clevr-Dist}}}
	& {{\rotatebox {90}{DMLab}}}
	& {{\rotatebox {90}{KITTI-Dist}}}
	& {{\rotatebox {90}{dSpr-Loc}}}
	& {{\rotatebox {90}{dSpr-Ori}}}
	& {{\rotatebox {90}{sNORB-Azim}}}
	&{{\rotatebox {90}{sNORB-Ele}}}
	& {{\rotatebox {90}{\textbf{Mean}}}}\\
 
	\specialrule{0em}{1pt}{1pt}
	\hline
	\specialrule{0em}{1pt}{1pt}

	\textcolor{gray}{Fully-FT} & \textcolor{gray}{85.8} & \textcolor{gray}{6.1} & \textcolor{gray}{89.21} & \textcolor{gray}{68.9} & \textcolor{gray}{87.7} & \textcolor{gray}{64.3} & \textcolor{gray}{97.2} & \textcolor{gray}{86.9} & \textcolor{gray}{87.4} & \textcolor{gray}{38.8} & \textcolor{gray}{79.7} & \textcolor{gray}{95.7} & \textcolor{gray}{84.2} & \textcolor{gray}{73.9} & \textcolor{gray}{56.3} & \textcolor{gray}{58.6} & \textcolor{gray}{41.7} & \textcolor{gray}{65.5} & \textcolor{gray}{57.5} & \textcolor{gray}{46.7} & \textcolor{gray}{25.7} & \textcolor{gray}{29.1} & \textcolor{gray}{68.9} \\
 
	\hline
	
	\specialrule{0em}{1pt}{1pt}
    
    BitFiT & \underline{0.10} & 3.8 & \underline{87.95} & 72.8 & 87.0 & 59.2 & 97.5 & 85.3 & 59.9 & 51.4 & 78.7 & 91.6 & 72.9 & 69.8 & 61.5 & 55.6 & 32.4 & 55.9 & 66.6 & 40.0 & 15.7 & 25.1 & 65.2 \\
    
    VPT & 0.53 & 5.6 & 72.10 & 78.8 & 90.8 & 65.8 & 98.0 & 88.3 & 78.1 & 49.6 & 81.8 & 96.1 & 83.4 & 68.4 & 68.5 & 60.0 & 46.5 & 72.8 & 73.6 & 47.9 & 32.9 & 37.8 & 72.0 \\
    
    RS-Bypass. & 0.42 & {2.4} & 47.53 & 64.5 & 88.8 & 73.2 & 99.4 & 90.6 & 63.5 & \underline{57.2} & 85.5 & 95.2 & 82.4 & 75.2 & 70.4 & 61.0 & 40.2 & 66.8 & 79.2 & 52.6 & 26.0 & \underline{49.3} & 72.3 \\
    
    Adapter & 0.16 & 3.9 & 82.78 & 69.2 & 90.1 & 68.0 & 98.8 & 89.9 & 82.8 & 54.3 & 84.0 & 94.9 & 81.9 & 75.5 & 80.9 & 65.3 & 48.6 & 78.3 & 74.8 & 48.5 & 29.9 & 41.6 & 73.9 \\
    
    LoRA & 0.29 & 3.9 & 85.03 & 67.1 & 91.4 & 69.4 & 98.8 & 90.4 & 85.3 & 54.0 & 84.9 & 95.3 & 84.4 & 73.6 & 82.9 & \textbf{69.2} & 49.8 & 78.5 & 75.7 & 47.1 & 31.0 & 44.0 & 74.5 \\
    
    AdaptFormer & 0.16 & 4.1 & 80.84 & 70.8 & 91.2 & 70.5 & 99.1 & 90.9 & 86.6 & 54.8 & 83.0 & 95.8 & 84.4 & \underline{76.3} & 81.9 & 64.3 & 49.3 & 80.3 & 76.3 & 45.7 & 31.7 & 41.1 & 74.7 \\
    
    NOAH & 0.36 & 5.2 & 56.78 & 69.6 & 92.7 & 70.2 & 99.1 & 90.4 & 86.1 & 53.7 & 84.4 & 95.4 & 83.9 & 75.8 & 82.8 & \underline{68.9} & 49.9 & \underline{81.7} & 81.8 & 48.3 & 32.8 & 44.2 & 75.5 \\
    
    FacT & 0.06 & 4.8 & -- & 70.6 & 90.6 & 70.8 & 99.1 & 90.7 & 88.6 & 54.1 & 84.8 & 96.2 & 84.5 & 75.7 & 82.6 & 68.2 & 49.8 & 80.7 & 80.8 & 47.4 & 33.2 & 43.0 & 75.6 \\
    
    Res-Tuning  & 0.55 & 4.9 & 57.70 & 75.2 & 92.7 & 71.9 & 99.3 & \textbf{91.9} & 86.7 & \textbf{58.5} & 86.7 & 95.6 & 85.0 & 74.6 & 80.2 & 63.6 & 50.6 & 80.2 & 85.4 & 55.7 & 31.9 & 42.0 & 76.3 \\
    
    Convpass & 0.33 & 2.6 & 67.02 & 72.3 & 91.2 & 72.2 & 99.2 & 90.9 & \underline{91.3} & 54.9 & 84.2 & 96.1 & 85.3 & 75.6 & 82.3 & 67.9 & 51.3 & 80.0 & 85.9 & 53.1 & 36.4 & 44.4 & 76.6 \\
    
    \hline
    HST  & 0.78 & 2.8 & 44.17 & 76.7 & \textbf{94.1} & 74.8 & \underline{99.6} & 91.1 & 91.2 & 52.3 & \textbf{87.1} & 96.3 & \textbf{88.6} & \textbf{76.5} & \textbf{85.4} & 63.7 & \textbf{52.9} & \underline{81.7} & \underline{87.2} & 56.8 & 35.8 & \textbf{52.1} & \underline{78.1} \\
    
    LoSA & \textbf{0.05} & \underline{2.1} & {68.92} & \textbf{82.7} & 93.0 & \textbf{76.2} & \textbf{99.7} & 89.8 & 80.0 & 56.1 & 86.3 & \underline{96.7} & 86.7 & \underline{76.3} & 78.8 & 61.4 & 48.0 & \textbf{82.6} & \textbf{91.7} & \textbf{58.4} & \textbf{46.9} & 47.6 & 77.8 \\

    UniPT & 5.19 & \underline{2.1} & 45.72 & 77.4 & 92.9 & 74.8 & \underline{99.6} & 91.6 & 89.7 & 54.9 & \underline{87.0} & 96.5 & 86.9 & 73.8 & 83.8 & 63.7 & 50.9 & 81.4 & 86.5 & 56.4 & 35.9 & 46.8 & 77.4\\

    SHERL & 5.08 & \textbf{2.0} & 48.70 & 78.8 & 93.2 & 75.4 & 99.5 & 91.6 & 90.4 & 55.6 & 86.8 & \underline{96.7} & 87.2 & 75.3 & 84.2 & 63.5 & 51.8 & \underline{81.7} & 86.9 & \underline{57.5} & 36.9 & 48.2 & 77.9 \\
    
    \textbf{Ours} & 5.50 & \underline{2.1} & \textbf{88.89} & \underline{79.8} & \underline{93.6} & \underline{75.7} & 99.5 & \underline{91.8} & \textbf{91.5} & 56.1 & 86.9 & \textbf{96.8} & \underline{87.5} & 75.9 & \underline{84.6} & 64.6 & \underline{52.4} & 81.3 & \underline{87.2} & {57.1} & \underline{37.7} & 47.9 & \textbf{78.3}\\

	\bottomrule[1.5pt]
	\end{tabular}
	\label{tab_vtab_results}
\end{table*}

\begin{table*}[!t]
    \caption{Comparison results with METL approaches across various architectures and distinct VL tasks. Ours${}^\heartsuit$ and Ours${}^\spadesuit$ denote our implementations based on UniPT and SHERL, respectively. Best results are in \textbf{bold}, and second best results are \underline{underlined}.
    }
    \vspace{-0.2cm}
    \begin{center}
    \setlength{\tabcolsep}{0.36pt}
    \resizebox{\textwidth}{!}{
    \begin{tabular}{l c c c ccc ccc ccc | c c c ccc ccc}
    \toprule[1.2pt]
    
    \multirow{2}{*}{Method} & Params. & \multicolumn{1}{c}{Mem.} & QPS & \multicolumn{3}{c}{Flickr30K} & \multicolumn{3}{c}{MSCOCO1K} & \multicolumn{3}{c|}{MSCOCO5K} & Params & \multicolumn{1}{c}{Mem.} & QPS & \multicolumn{3}{c}{MSR-VTT} & \multicolumn{3}{c}{MSVD} \\
    
    \cmidrule[0.4pt]{5-13}
    \cmidrule[0.4pt]{17-22}
    & (M) $\!\downarrow$ & (G) $\!\downarrow$ & $\!\uparrow$ & I-T $\!\uparrow\ $ & T-I $\!\uparrow\ $ & Rsum $\!\uparrow\ $ & I-T $\!\uparrow\ $ & T-I $\!\uparrow\ $ & Rsum $\!\uparrow\ $ & I-T $\!\uparrow\ $ & T-I $\!\uparrow\ $ & Rsum $\!\uparrow\ $ & (M) $\!\downarrow$ &  (G) $\!\downarrow$ & $\!\uparrow$ & T-V $\!\uparrow\ $ & V-T $\!\uparrow\ $ & Rsum $\!\uparrow\ $ & T-V $\!\uparrow\ $ & V-T $\!\uparrow\ $ & Rsum $\!\uparrow\ $ \\
    
    \midrule[0.6pt]
    \textcolor{gray}{Fully-FT} & \textcolor{gray}{201.2} & \textcolor{gray}{176.8} & \textcolor{gray}{-} & \textcolor{gray}{85.6} & \textcolor{gray}{73.3} & \textcolor{gray}{546.6} & \textcolor{gray}{83.1} & \textcolor{gray}{71.7} & \textcolor{gray}{542.7} & \textcolor{gray}{64.2} & \textcolor{gray}{51.2} & \textcolor{gray}{468.9} & \textcolor{gray}{151.3} & \textcolor{gray}{48.8} & \textcolor{gray}{-} & \textcolor{gray}{42.8} & \textcolor{gray}{42.1} & \textcolor{gray}{389.2} & \textcolor{gray}{45.2} & \textcolor{gray}{57.1} & \textcolor{gray}{425.5}  \\
    
    \midrule[0.6pt]
    LST \cite{sung2022lst} & \textbf{9.7} & \textbf{24.4} & - & 82.1 & 66.5 & 529.5 & 78.2 & 64.8 & 525.8 & 57.8 & 43.1 & 434.5 & 11.2 & 32.0 & - & 37.0 & 37.8 & 356.7 & 35.5 & 55.4 & 407.2\\
    UniPT \cite{liao2024uni} & 12.4 & \textbf{24.4} & 8.61 & 84.8 & 69.1 & 537.4 & 80.6 & 67.5 & 532.9 & 61.1 & 45.9 & 445.3 & \textbf{9.6} & \textbf{13.6} & 27.03 & 38.9 & 39.3 & 361.3 & 40.9 & 59.7 & 432.1\\
    
    SHERL \cite{diao2024sherl} & \underline{11.3} & \textbf{24.4} & 10.24 & \underline{86.1} & \underline{71.1} & \underline{542.3} & \underline{81.8} & \textbf{69.2} & \underline{537.5} & \underline{62.5} & \underline{47.3} & \underline{450.8} & \textbf{9.6} & \textbf{13.6} & 32.53 & 39.2 & 40.6 & 363.7 & 40.9 & \underline{60.2} & 429.7\\
    
    \textbf{Ours${}^\heartsuit$} & 12.4 & \underline{25.1} & \underline{14.71} & 85.9 & 70.7 & 541.2 & 81.3 & \underline{68.8} & 536.4 & 61.9 & 46.8 & 448.1 & \underline{9.7} & \underline{14.0} & \textbf{43.65} & \underline{40.2} & \underline{40.7} & \underline{365.8} & \underline{42.3} & 60.0 & \underline{435.7}\\
    
    \textbf{Ours${}^\spadesuit$} & 11.4 & {25.2} & \textbf{14.83} & \textbf{86.4} & \textbf{71.6} & \textbf{543.0} & \textbf{82.2} & \textbf{69.2} & \textbf{537.9} & \textbf{63.2} & \textbf{48.0} & \textbf{452.3}& \underline{9.7} & 14.2 & \underline{43.52} & \textbf{40.6} & \textbf{41.2} & \textbf{367.2} & \textbf{42.5} & \textbf{61.3} & \textbf{435.8}\\
    \bottomrule[1.2pt]
    \end{tabular}
    }
    \resizebox{\textwidth}{!}{
    \begin{tabular}{l c c c cc cc | c c c c ccc ccc cc}
    \multirow{2}{*}{Method} & Params. & \multicolumn{1}{c}{Mem.} & Time & \multicolumn{2}{c}{VQAv2} & \multicolumn{2}{c|}{GQA} 
    & Params. & \multicolumn{1}{c}{Mem.} & Time & \multicolumn{3}{c}{RefCOCO} & \multicolumn{3}{c}{RefCOCO+} & \multicolumn{2}{c}{RefCOCOg}\\
    
    \cmidrule[0.4pt]{5-8}
    \cmidrule[0.4pt]{12-19}
    & (M) $\!\downarrow$ & (G) $\!\downarrow$ & (ms) $\!\downarrow\ $ & Test$_\text{D}$ $\!\uparrow$ & Test$_\text{S}$ $\!\uparrow$ & Test$_\text{D}$ $\!\uparrow$ & Test$_\text{S}$ $\!\uparrow\ $ & (M) $\!\downarrow$ & (G) $\!\downarrow$ & (ms) $\!\downarrow\ $ & Val $\!\uparrow$ & TestA $\!\uparrow$ & TestB $\!\uparrow$ & Val $\!\uparrow$ & TestA $\!\uparrow$ & TestB $\!\uparrow$ & Val $\!\uparrow$ & Test $\!\uparrow$\\
    
    \midrule[0.6pt]
    \textcolor{gray}{Fully-FT} & \textcolor{gray}{236.8} & \textcolor{gray}{82.0} & \textcolor{gray}{-} & \textcolor{gray}{76.71} & \textcolor{gray}{76.86} & \textcolor{gray}{60.25} & \textcolor{gray}{61.44}& \textcolor{gray}{185.2} & \textcolor{gray}{39.6} & \textcolor{gray}{-} & \textcolor{gray}{86.51} & \textcolor{gray}{89.13} & \textcolor{gray}{81.22} & \textcolor{gray}{79.54} & \textcolor{gray}{84.54} & \textcolor{gray}{70.63} & \textcolor{gray}{80.92} & \textcolor{gray}{80.95} \\
    \midrule[0.6pt]
    LST \cite{sung2022lst} & 13.4 & 25.6 & - & 75.29 & 75.44 & 59.93 & 60.75 & 0.9 & 12.6 & - & 81.63 & 85.19 & 76.03 & 71.32 & 78.20 & 62.06 & 72.53 & 73.67\\
    UniPT \cite{liao2024uni} & \textbf{10.3} & \textbf{11.6} & 106.38 & 75.33 & 75.53 & 60.10 & 60.72 & \textbf{0.7} & \textbf{6.8} & 97.82 & 82.71 & 86.25 & 78.16 & 72.94 & 79.18 & 64.49 & 77.04 & 77.33\\
    SHERL \cite{diao2024sherl} & 13.0 & 14.0 & 92.49 & 75.53 & 75.82 & 60.16 & 60.82 & \textbf{0.7} & \textbf{6.8} & 79.33 & 83.02 & 86.39 & 78.41 & 73.29 & 80.11 & 64.59 & \underline{77.80} & 77.33\\
    
    \textbf{Ours${}^\heartsuit$} & \underline{10.4} & \underline{12.2} & \underline{60.26} & \underline{75.57} & \underline{75.91} & \underline{60.21} & \underline{60.85} & \textbf{0.7} & \underline{7.1} & \underline{52.63} & \textbf{83.11} & \underline{86.64} & \underline{78.69} & \underline{73.49} & \underline{80.15} & \underline{64.72} & 77.49 & \underline{77.73}\\
    
    \textbf{Ours${}^\spadesuit$} & 13.1 & 15.2 & \textbf{59.89} & \textbf{75.88} & \textbf{76.07} & \textbf{60.41} & \textbf{60.93} & \underline{0.8} & \underline{7.1} & \textbf{52.29} & \underline{83.09} & \textbf{86.77} & \textbf{78.97} & \textbf{74.05} & \textbf{80.46} & \textbf{64.79} &\textbf{78.06} & \textbf{78.13}\\
    \bottomrule[1.2pt]
    \end{tabular}
    }
    \end{center}
\vspace{-0.3cm}
        \label{METL-table}
\end{table*}

\begin{table*}[!t]
     \caption{Ablation study on various distillation strategies that are applied to distinct structures (\emph{i.e.} `logits-based' and `feature-based') with different distillation directions, where `S' and `B' refer to side network and backbone, respectively. Best results are in \textbf{bold}, and the second best ones are \underline{underlined}.
    }
    \centering
    \begin{tabular}{cccccccccc}
         \toprule[1.2pt]
         \multirow{2}{*}{Method} & {Params.} & {Memory} & {QPS} & \multicolumn{3}{c}{Flickr30K} & \multicolumn{3}{c}{MSCOCO1K} \\
         \cmidrule[0.4pt]{5-10}
         & (M) $\downarrow$ & (G) $\downarrow$ & $\uparrow$ & {I-T} $\uparrow$ & {T-I} $\uparrow$ & {Rsum} $\uparrow$ & {I-T} $\uparrow$ & {T-I} $\uparrow$ & {Rsum} $\uparrow$\\
         \midrule[0.6pt]
         Baseline (UniPT) \cite{liao2024uni} & \textbf{12.4} & \textbf{24.4} & \underline{8.61} & \underline{84.8} & 69.1 & 537.4 & \textbf{80.6} & \underline{67.5} & \textbf{532.9} \\
         \midrule[0.6pt]
         {Logits-based (S$\to$B)} & \textbf{12.4} & \underline{24.7} & \textbf{14.71} & 81.2 & 66.1 & 529.7 & 77.2 & 65.0 & 526.2 \\
         \midrule[0.3pt]
         {Feature-based (B$\to$S)} & \textbf{12.4} & \underline{24.7} & \textbf{14.71} & 81.9 & 66.5 & 531.5 & 78.3 & 65.9 & 528.1 \\
         \midrule[0.3pt]
         {Feature-based (S$\to$B)} & \textbf{12.4} & \underline{24.7} & \textbf{14.71} & 78.9 & 63.0 & 518.2 & 73.1 & 60.9 & 508.3 \\
         \midrule[0.3pt]
         \textbf{Ours} (B $\leftrightarrows$ S) & \textbf{12.4} & 25.1& \textbf{14.71} & \textbf{84.9} & \textbf{69.5} & \textbf{538.1} & 80.2 & \textbf{67.7} & \underline{532.7} \\
         \bottomrule[1.2pt]
    \end{tabular}
    \label{tab_ablation_distillation}
\end{table*}

\begin{table*}[!t]
     \caption{Ablation study on fading the side network during inference on distinct datasets for the ITR task. Best results are in \textbf{bold}, and the second best ones are \underline{underlined}.}
    \centering
    \begin{tabular}{ccccccccccc}
         \toprule[1.2pt]
         \multirow{2}{*}{Method} & Fading & {Params.} & {Memory} & {QPS} & \multicolumn{3}{c}{Flickr30K} & \multicolumn{3}{c}{MSCOCO1K} \\
         \cmidrule[0.4pt]{6-11}
         & Side Network & (M) $\downarrow$ & (G) $\downarrow$ & $\uparrow$ & {I-T} $\uparrow$ & {T-I} $\uparrow$ & {Rsum} $\uparrow$ & {I-T} $\uparrow$ & {T-I} $\uparrow$ & {Rsum} $\uparrow$\\
         \midrule[0.6pt]
         \textcolor{gray}{Fully-FT} & \textcolor{gray}{$\times$} & \textcolor{gray}{201.2} & \textcolor{gray}{176.8} & \textcolor{gray}{-} & \textcolor{gray}{85.6} & \textcolor{gray}{73.3} & \textcolor{gray}{546.6} & \textcolor{gray}{83.1} & \textcolor{gray}{71.7} & \textcolor{gray}{542.7} \\
         LST \cite{sung2022lst} & $\times$ & \textbf{9.7} & \textbf{24.4} & - & 82.1 & 66.5 & 529.5 & 78.2 & 64.8 & 525.8 \\ 
         \midrule[0.6pt]
         UniPT \cite{liao2024uni} & $\times$ & 12.4 & \textbf{24.4} & 8.61 & 84.8 & 69.1 & 537.4 & 80.6 & 67.5 & 532.9\\

         \multirow{2}{*}{\textbf{\textbf{Ours${}^\heartsuit$}}} & $\times$ & 12.4 & \underline{25.1} & 7.98 & 86.2 & 70.9 & 541.9 & 81.8 & \underline{69.4} & 537.6\\

         & \checkmark & 12.4 & \underline{25.1} & \underline{14.71} & 85.9 & 70.7 & 541.2 & 81.3 & {68.8} & 536.4\\
         \midrule[0.6pt]
         SHERL \cite{diao2024sherl} & $\times$ & \underline{11.3} & \textbf{24.4} & 10.24 & {86.1} & {71.1} & \underline{542.3} & \underline{81.8} & {69.2} & {537.5}\\

         \multirow{2}{*}{\textbf{Ours${}^\spadesuit$}} & $\times$ & 11.4 & 25.2 & 9.76 & \textbf{87.0} & \underline{71.4} & \textbf{543.9} & \textbf{82.5} & \textbf{69.8} & \textbf{539.0}\\
         
         & \checkmark & 11.4 & {25.2} & \textbf{14.83} & \underline{86.4} & \textbf{71.6} & \underline{543.0} & \underline{82.2} & {69.2} & \underline{537.9}\\
         \bottomrule[1.2pt]
    \end{tabular}
    \label{tab_ablation_fading}
\end{table*}

\subsection{Main Results}
Following \cite{diao2024unipt}, we compare MDPD against state-of-the-art PETL and METL methods, while Fully Fine-Tuning (Fully-FT) and Partially tuning baselines without comparison. 
\subsubsection{Our Method vs. PETL Approaches}
{We compare to the following approaches without adopting memory reduction strategies}: 1) Selective Tuning methods including BitFit \cite{zaken2022bitfit}, and FacT \cite{jie2023fact}; 2) Additive Tuning methods including Prompt \cite{li2021prefix}, Adapter \cite{houlsby2019param}, AdaptFormer \cite{chen2022adaptformer}, NOAH \cite{zhang2024neural}, Convpass \cite{jie2024convolutional}, Res-Tuning \cite{jiang2024res}; 3) Re-parameterized Tuning methods including LoRA \cite{hu2022lora}, SSF \cite{lian2022scaling}, AdaLoRA \cite{zhang2023adaptive}. On VTAB-1K benchmark, we additionally compare with VPT \cite{jia2022visual}.

\textbf{Results on VL Tasks.} 
We employ two structures from \textit{VSE$\infty$} \cite{chen2021VSE} with BERT-base+BUTD regions and ResNeXt-101+BiGRU for ITR, and one structure from \textit{CLIP4Clip} \cite{luo2021clip4clip} with ViT-base+Text Transformer for VTR. As shown in Table~\ref{tab1}, our method promotes accuracy of compared PETL approaches in most cases. Despite adopting slightly more learnable parameters, our method substantially reduces training memory overhead, by least 59.5\%, 90.4\% and 64.3\% with various backbones respectively, through optimizing lightweight side networks instead of backbones. It also yields the highest retrieval rate and the fastest inference time by abandoning side networks after training.


\textbf{Results on Language-only Tasks.}
Table~\ref{tab_glue_results} shows the performance on GLUE with T5-base \cite{raffel2020exploring} by default. Our method decreases training memory by at least 68.2\% and reaches the minimal inference latency compared to PEFT approaches, while maintaining competitive accuracy with 
comparable trainable parameters. When incorporating T5-large, our method promotes accuracy of PETL methods by a large margin with lower memory usage. 


\textbf{Results on Vision-only Tasks.}
We evaluate the performance of our method for vision-only tasks on VTAB-1K benchmark by adopting a ViT-B pre-trained on ImageNet-21K. As summarized in Table~\ref{tab_vtab_results}, our method achieves a new SOTA performance on average across 19 datasets. Benefiting from side network with dual path knowledge distillation, our method is clearly more memory efficient than compared PETL approaches, reducing the training memory usage, while maintaining the least inference time cost.

\subsubsection{Our Method vs. METL Approaches}
We further compare with SOTA METL approaches, including LST \cite{sung2022lst}, UniPT \cite{diao2024unipt}, SHERL \cite{diao2024sherl}, HST \cite{lin2023hierarchical} and LoSA \cite{mercea2024time}. It is worth noting that not all compared methods are fully evaluated on the selected benchmarks. Particularly, UniPT and SHERL did not report results on VTAB-1K. We therefore reproduce their results by using the released source code, while adopting the reported results for the others. As shown in Tables~\ref{tab1}-\ref{tab_vtab_results}, our method consistently promotes accuracy of existing METL methods through mutually distilling backbones and side networks across multiple layers with fine-grained distillation strategies. For instance, in Table~\ref{tab1}, we suppresses LST/UniPT by average gains of 3.4\%/1.4\% in R@1 and 11.9\%/5.7\% in Rsum on cross-model retrieval.

Additionally, we expand evaluations for VL tasks on more downstream datasets and network architectures including VSE$\infty$~\cite{chen2021VSE}, CLIP4Clip~\cite{luo2021clip4clip}, CLIP-ViL~\cite{shen2021much} and MDETR~\cite{kamath2021mdetr}, of which details are depicted in \emph{Supplementary Material}. As shown in Table~\ref{METL-table}, MDPD clearly promotes accuracy of compared METL method across various VL tasks by maintaining almost same training memory overhead. Particularly, based on UniPT/SHERL, our method achieves an average improvement of 1.0/0.7\% in R@1 and 3.6/2.4\% in Rsum for cross-modal retrieval, about 0.86/0.96\% improvements on QA task, and total benefits of 3.92/3.38\% for VG task, respectively. Moreover, benefiting from the design of our side branch that is utilized only for training and is abandoned during inference, our method remarkably accelerates the inference speed. Notably, our method decreases the inference time cost of SHERL by at least 25.2\%, clearly showing the efficiency of our method.

\subsection{Ablation Study}

\textbf{Effect of Dual Path Knowledge Distillation.}
Our method introduces a dual-path distillation framework that integrates both feature-based distillation from backbone to side network and logits-based distillation from side network to backbone. To validate the effectiveness of the approach, we compare different distillation strategies. As shown in Table~\ref{tab_ablation_distillation}, the combination of logits- and feature-based distillation consistently outperforms compared strategies, highlighting their complementary benefits. Additionally, utilizing a bottleneck for feature-based distillation significantly reduces the amount of learnable parameters and training memory without sacrificing the accuracy. The dual path distillation, which combines both logits- and feature-based distillation, proves essential for achieving SOTA performance while maintaining computational efficiency.

\textbf{Effect of Fading the Side Network in Inference.}
To further clarify the influence of fading the side network on performance during inference, we conduct ablation study by comparing with state-of-the-art METL approaches. As shown in Table~\ref{tab_ablation_fading}, when adopting both the backbone and side network, our method achieves improvements of 13.0\%, 4.6\% and 1.6\% in Rsum compared to LST \cite{sung2022lst}, UniPT \cite{liao2024uni} and SHERL \cite{diao2024sherl}, respectively, under comparable efficiency including the number of trainable parameters, memory usage and inference time. Furthermore, by discarding the side network during inference, our method significantly gains at least 44.8\% improvement in queries per second, with a slight performance decrease of about 0.9\%. The results clearly demonstrate the effectiveness of fading the side branch, when applying our proposed approach.

We also extensively study the hierarchical feature-based distillation module, including \textbf{influence of the mask ratio $\bm{\lambda}$} in Eq.~\eqref{eq:mask}, \textbf{effect of generation block $\bm{\mathcal{G}}$} in Eq.~\eqref{eq:generation_block} and \textbf{impact of feature distillation loss} on distinct layers. Due to space limitation, we summarize the detailed results in \emph{Supplementary Material}.

%% file: main_sec/5_conclusion.tex
\section{Conclusion}
In this paper, we propose a novel approach dubbed Masked Dual Path Distillation (MDPD) to accelerate the memory-efficient transfer learning based on side networks. Specifically, MDPD leverages the features of side networks to efficiently guide the fine-tuning of a small number of parameters of backbone during training, and abandons side networks during inference. To maintain the accuracy, we propose a Hierarchical Feature-based Distillation strategy to enhance the optimization of backbones. We extensively evaluate the performance of MDPD on various pre-trained models across the vision-language and vision/language-only tasks. The results reveal that our method remarkably reduce the inference time, while maintaining the accuracy as well as the memory and parameter efficiency.

%% file: main_sec/6_acknowledgment.tex
\section*{Acknowledgments}
\label{sec:acknowledgments}
{This work was partly supported by the Beijing Natural Science Foundation (No. 4242044 and No. L259044), the CCF Baidu Open Fund, and the Fundamental Research Funds for the Central Universities.}

%% file: main_sec/X_suppl.tex

\clearpage
\setcounter{page}{1}
\setcounter{section}{0}
\renewcommand\thesection{\Alph{section}}
\setcounter{figure}{0}
\setcounter{table}{0}
\renewcommand{\thefigure}{\Alph{figure}}
\renewcommand{\thetable}{\Alph{table}}

\maketitlesupplementary


In the Supplementary Material, more details and experiments are organized as follows, we additionally provide more introduction of \textbf{Datasets and Metrics} in~\cref{sec:supple_datasets}, the \textbf{Implementation Details} in~\cref{sec:supple_implementation}, the rigorous proof of \textbf{Back-propagation through Large Backbone} in~\cref{sec:supple_BP}, the expanded explanation of \textbf{Baselines} in~\cref{sec:supple_baselines}, and \textbf{More Ablation Studies} on several modules in~\cref{sec:supple_ablation}.


\section{Datasets and Metrics}
\label{sec:supple_datasets}
\textbf{Image-Text Retrieval (ITR)}: We employ Flickr30k \cite{young2014image} and MSCOCO \cite{lin2014microsoft} for image-text matching task. The Flickr30k dataset contains 31,783 colloquial images with 158,915 captions, partitioned into 29,783 training images, 1,000 validation images, and 1,000 test images. Its captions emphasize explicit compositional semantics (\emph{e.g.}, object-attribute-spatial relationships), posing fine-grained matching challenges. The MSCOCO dataset comprises 123,287 complex scene images annotated with 616,435 captions, and is divided into 113,287 training images, 5,000 validation images, 5,000 test images, which is the standard Karpathy split. MSCOCO requires deeper relational reasoning due to contextual object interactions and scene dynamics. Both datasets evaluate Image-to-Text (I-T), Text-to-Image (T-I) retrieval using Recall@1, with additionally employing RSum (\emph{i.e.}, sum of all six Recall@K scores where K=1,5,10) as the holistic metric reflecting their compositional complexity.

\textbf{Video-Text Retrieval (VTR)}: For video-text retrieval, we employ MSR-VTT \cite{xu2016msr} and MSVD \cite{chen2011collecting}. The MSR-VTT (\emph{i.e.}, Microsoft Research Video to Text) dataset contains 10,000 web video clips (total duration ~41 hours) sourced from YouTube, with each video annotated with 20 English captions. Following the standard split, we employ the 1k-A protocol, where 9,000 videos with all corresponding captions for training, and 1,000 pairs are for testing. Characterized by high diversity across 20 categories (\emph{e.g.}, sports, music, news), its captions describe complex temporal dynamics and object interactions. Moreover, The MSVD (\emph{i.e.}, Microsoft Video Description) dataset comprises 1,970 short video clips with approximately 80,000 multilingual captions, partitioned into 1,200 training videos, 100 validation videos, and 670 testing videos. Noted for fine-grained temporal alignment challenges, MSVD captions emphasize precise action-object localization. Both datasets evaluate Video-to-Text (V-T), Text-to-Video (T-V) retrieval using Recall@1 and RSum for evaluation.

\textbf{Question Answering (VQA\&GQA)}: For question answering tasks, we evaluate on VQAv2 \cite{goyal2017making} and GQA \cite{hudson2019gqa}. The VQAv2 dataset addresses prior language bias issues by pairing 204,721 COCO images with 1,105,904 questions. It employs the standard split: 82,783 images with 443,757 questions for training, 40,504 images with 214,354 questions for validation, and 81,434 images with 447,793 questions for testing. Questions require diverse reasoning about object attributes, actions, and scene context. The GQA dataset features 113,018 images with 22,669,678 questions generated from scene graphs to ensure compositional rigor. Its balanced split contains 943,000 training questions ($\sim$70\%), 132,062 validation questions ($\sim$10\%)), and 264,159 testing questions ($\sim$20\%), emphasizing structural reasoning over 1,704 object categories. We evaluate performance on both Test-Dev and Test-Std splits via the official EvalAI system.

\textbf{Visual Grounding (VG)}: We utilize the RefCOCO, RefCOCO+ \cite{yu2016modeling}, and RefCOCOg \cite{mao2016generation} derived from MSCOCO images for visual grounding. RefCOCO contains 19,994 images with 50,000 bounding boxes annotated by 142,210 expressions, whose standard UNC split comprises 120,624 training expressions, 10,834 validation expressions, and 5,675/5,095 Test A/B expressions. Test A focuses on bounding boxes containing person instances, while Test B involves non-person objects. RefCOCO+ shares the same image set but introduces stricter constraints: 49,856 referred objects with 141,564 expressions that explicitly prohibit location words, divided into 120,191 training, 10,758 validation, and 5,726/4,889 Test A/B expressions. RefCOCOg differs substantially with 26,711 images, 54,822 referred objects, and 104,560 longer, grammatically complex expressions, while they are categorized into train, validation, and test, with 85,474, 7,323, and 9,592 samples. Primary evaluation uses precision@0.5 to measure localization accuracy of predicted bounding boxes against human annotations.

{\begin{table*}[!ht]
    \centering
    \caption{Detailed Hyper-parameters of MDPD on ITR, VTR, VQA, GQA, and VG tasks. Among them, \textit{AdamW} is adopted as the optimizer uniformly.}  
    \vskip 0.1in
    \begin{center}
    {\begin{tabular}{lcccccc}
         \toprule[1.2pt]
         Task & Model & Learning Rate & Optimizer ($\beta_1$, $\beta_2$, Weight Decay) & Batch Size & Total Epochs & Warmup Strategy \\
         \midrule[0.6pt]
         ITR & VSE$\infty$ & $5 \times 10^{-4}$ & $0.9, 0.999, 1 \times 10^{-2}$ & $112$ & $25$ & linear \\
         VTR & CLIP4Clip & $1 \times 10^{-4}$ & $0.9, 0.98, 1 \times 10^{-2}$ & $128$ & $5$ & cosine \\
         VQA & CLIP-ViL & $5 \times 10^{-4}$ & $0.9, 0.999, 1 \times 10^{-2}$ & $256$ & $5$ & linear \\
         GQA & CLIP-ViL & $1 \times 10^{-4}$ & $0.9, 0.999, 1 \times 10^{-2}$ & $256$ & $5$ & linear \\
         VG & MDETR & $5 \times 10^{-4}$ & $0.9, 0.999, 0$ & $8$ & $10$ & linear \\
         \bottomrule[1.2pt]
    \end{tabular}}
    \end{center}
    \label{tab_vl_hyperparameters}
\end{table*}}

\textbf{Language-only}: For language-only task, we employ the General Language Understanding Evaluation (GLUE) benchmark \cite{wang2018glue} consolidates eight NLP tasks into four core categories, consisting of \textit{linguistic acceptability} (CoLA \cite{warstadt2019neural}), \textit{sentiment analysis} (SST-2 \cite{socher2013recursive}), \textit{similarity and paraphrase} (MRPC \cite{dolan2005automatically}, QQP, STS-B \cite{cer2017semeval}), and \textit{natural language inference} (MNLI \cite{williams2017broad}, QNLI \cite{rajpurkar2016squad}, RTE \cite{bentivogli2009fifth}). Evaluation employs task-specific metrics: classification Accuracy metric for SST-2, MNLI, RTE, and QNLI; F1-score augmented with accuracy for MRPC and QQP; Matthew's Correlation for the class-imbalanced data of CoLA; and Pearson-Spearman Correlation for similarity scoring of STS-B.

\textbf{Vision-only}: For vision-only task, we employ the Visual Task Adaptation Benchmark (VTAB-1K) \cite{zhai2019large} systematically evaluates transfer learning capabilities across 19 diverse vision datasets unified under a standardized low-data regime, categorized into three distinct task types: (1) \textit{Natural} tasks (CIFAR-100 \cite{krizhevsky2009learning}, Caltech101 \cite{fei2006one}, DTD \cite{cimpoi2014describing}, Flowers102 \cite{nilsback2008automated}, Pets \cite{parkhi2012cats}, SVHN \cite{netzer2011reading}, Sun397 \cite{xiao2010sun}) featuring object-centric photographs with moderate complexity; (2) \textit{Specialized} tasks (Patch Camelyon \cite{veeling2018rotation}, EuroSAT \cite{helber2019eurosat}, Resisc45 \cite{cheng2017remote}, Retinopathy) \cite{graham2015kaggle} comprising domain-specific imagery and medical images; (3) \textit{Structured} tasks (Clevr/count \cite{johnson2017clevr}, Clevr/distance \cite{johnson2017clevr}, DMLab \cite{beattie2016deepmind}, KITTI-Dist \cite{geiger2013vision}, dSprites/location, dSprites/orientation, SmallNORB/azimuth \cite{lecun2004learning}, SmallNORB/elevation \cite{lecun2004learning}) emphasizing geometric relationships and spatial reasoning. Each dataset provides 1,000 training images with predefined validation/test splits. Evaluation reports Top-1 Accuracy metric for classification tasks, with final performance aggregated via uniform averaging across all 19 datasets.


\section{Implementation Details}
\label{sec:supple_implementation}
For vision-language (VL) tasks, Table \ref{tab_vl_hyperparameters} comprehensively details the hyper-parameter configurations. More specifically, for ITR task using VSE$\infty$, we set the batch size to 112, and maintain consistency with pre-trained models for all other tasks, while scale learning rate by a factor of 10, and set the reduction factor and mask rate $\lambda$ to 2 and 0.5, respectively. For GLUE benchmark evaluations, we set learning rate, reduction factor and batch size to $3 \times 10^{-3}$, 8, and 100, respectively. Consistent with the methodology in LST~\cite{sung2022lst}, we implement the layer-dropping strategy: for the T5-base architecture, this entails removal of the $0^{\text{th}}$, $4^{\text{th}}$, and $8^{\text{th}}$ encoder/decoder layers; for T5-large, we omit all even-indexed encoder and decoder layers.

Beyond task-specific hyper-parameters, we carefully calibrate the weighting coefficients for multi-objective optimization. Specifically, we assign the logits-based and feature-based distillation loss in deep or shallow layers with weights of $1 \times 10^{-4}$, $6 \times 10^{-5}$ and $4 \times 10^{-5}$, respectively, while 1 for the primary Supervised Fine-Tuning objective. This balanced scheme ensures commensurate contribution from each optimization component during gradient updates. All experiments are conducted on an NVIDIA GeForce RTX 3090Ti GPU.

\section{Back-propagation through Large Backbone}
\label{sec:supple_BP}
Consider a neural network comprising $L$ sequential layers, where the transformation at the $i^{\text{th}}$ layer is defined as $f_i(\mathbf{x}) = \sigma_i(\mathbf{W}_i \mathbf{x} + \mathbf{b}_i)$. This composite function depends on the previous layer's output, parameterized by the weight matrix $\mathbf{W}_i$, bias vector $\mathbf{b}_i$, and nonlinear activation function $\sigma_i(\cdot)$. We denote the pre-activation output as $\mathbf{z}_{i+1}$ and the post-activation output as $\mathbf{a}_{i+1}$, establishing the layer-wise propagation:

\begin{equation}
    \mathbf{a}_{i+1} = \sigma_i(\mathbf{z}_{i+1}) = \sigma_i(\mathbf{W}_i \mathbf{a}_i + \mathbf{b}_i).
\end{equation}

Network parameters are optimized via stochastic gradient descent (SGD) by minimizing a scalar loss function $\mathcal{L}$ applied to the final layer output. The backpropagation algorithm computes gradients for $\mathbf{W}_i$ and $\mathbf{b}_i$ through recursive application of the multivariate chain rule:

\begin{equation}
\label{wb_bp}
    \begin{aligned}
        \frac{\partial \mathcal{L}}{\partial \mathbf{W}_i} 
        &= \frac{\partial \mathcal{L}}{\partial \mathbf{a}_{i+1}} 
           \frac{\partial \mathbf{a}_{i+1}}{\partial \mathbf{z}_{i+1}}
           \frac{\partial \mathbf{z}_{i+1}}{\partial \mathbf{W}_i} 
        = \frac{\partial \mathcal{L}}{\partial \mathbf{a}_{i+1}} \sigma_i' \mathbf{a}_i^\top, \\
        \frac{\partial \mathcal{L}}{\partial \mathbf{b}_i} 
        &= \frac{\partial \mathcal{L}}{\partial \mathbf{a}_{i+1}} 
           \frac{\partial \mathbf{a}_{i+1}}{\partial \mathbf{z}_{i+1}} 
        = \frac{\partial \mathcal{L}}{\partial \mathbf{a}_{i+1}} \sigma_i',
    \end{aligned}
\end{equation}
where $\sigma_i' \equiv \frac{d\sigma_i}{d\mathbf{z}_{i+1}}$ denotes the activation gradient, and $\frac{\partial \mathcal{L}}{\partial \mathbf{a}_{i+1}}$ represents the upstream gradient from subsequent layers. This upstream gradient is recursively computed via backward propagation from layer $i+2$:

\begin{equation}
\label{a_bp}
    \frac{\partial \mathcal{L}}{\partial \mathbf{a}_{i+1}} 
    = \frac{\partial \mathcal{L}}{\partial \mathbf{a}_{i+2}} 
      \frac{\partial \mathbf{a}_{i+2}}{\partial \mathbf{z}_{i+2}} 
      \frac{\partial \mathbf{z}_{i+2}}{\partial \mathbf{a}_{i+1}} 
    = \frac{\partial \mathcal{L}}{\partial \mathbf{a}_{i+2}} \sigma_{i+1}' \mathbf{W}_{i+1}^\top.
\end{equation}

As formalized in Equations \eqref{wb_bp} and \eqref{a_bp}, the backpropagation algorithm incurs substantial computational overhead due to floating-point operations (FLOPs) required for two critical gradient components: 1) The activation gradients $\{\mathbf{a}\}$ corresponding to updated parameters $\{\mathbf{W}\}$, and 2) The activation derivatives $\{\sigma'\}$ that must be cached throughout the computational graph, where $\{\cdot\}$ denotes sets of activations, parameters, or gradients. Existing Parameter-Efficient Transfer Learning (PETL) techniques, including Adapter~\cite{houlsby2019param}, Prompt-Tuning~\cite{lester2021power}, and LoRA~\cite{hu2022lora}, mitigate memory footprint by reducing the parameter update set $|\{\mathbf{W}\}|$ through learning only sparse parameter subsets, where $|\{\cdot\}|$ means the size of set $\{\cdot\}$. Consequently, the memory allocated for activation storage $|\{\mathbf{a}\}|$ proportionally decreases. However, the dominant computational burden during backpropagation stems from computing gradient terms involving $\{\sigma'\}$ - the derivatives of activation functions. Crucially, $|\{\sigma'\}|$ remains undiminished in these methods since:
\begin{equation}
|\{\sigma'\}| = \sum_{i=1}^{L} \dim(\mathbf{z}_i).
\end{equation}

This persistence occurs because PETL methods typically introduce trainable parameters into network inputs or intermediate structures while keeping the backbone frozen. Nevertheless, they still require full computation of $\sigma'$ across all backbone operations, necessitating: 1) Complete evaluation of activation gradients through the entire computational graph, 2) Storage of intermediate derivatives at each layer, and 3) Backpropagation through all nonlinear transformations.

Since activation dimensions generally satisfy $|\{\mathbf{a}\}| = |\{\sigma'\}|$ (barring dimensionality-altering activations), the theoretical memory reduction ceiling becomes:
\begin{equation}
\text{Memory}_{\text{BP}} = \underbrace{|\{\mathbf{a}\}|}_{\text{reduced}} + \underbrace{|\{\sigma'\}|}_{\text{unchanged}} \leq 50\% \text{ reduction}.
\end{equation}

Therefore, while PETL methods reduce parameter update costs, they still incur substantial FLOPs and memory requirements proportional to backbone complexity, as full error backpropagation through frozen layers remains mandatory.

Based on the foregoing computational analysis, \textbf{side network} is proposed as a memory-efficient alternative. This lightweight network maintains same structure to the backbone network while scaling all weight matrices and hidden state dimensions by a reduction factor $r \geq 2$. Thus, the original backpropagation memory footprint $|\{\mathbf{a}\}| + |\{\sigma'\}|$ is fundamentally transformed in this paradigm. Crucially, the side network \textit{decouples} from the backbone's computational graph during backpropagation, requiring gradient computation only through its own structure. Consequently, its memory consumption reduces to:
\begin{equation}
\text{Memory}_{\text{BP}}^{\text{side}} = \frac{|\{\mathbf{a}\}| + |\{\sigma'\}|}{r}
\end{equation}

This yields a critical comparative advantage: when $r > 2$, the side network achieves strictly lower memory consumption than the theoretical minimum of Parameter-Efficient Transfer Learning (PETL) methods, which remain bounded by:
\begin{equation}
\text{Memory}_{\text{BP}}^{\text{PETL}} \geq \frac{|\{\mathbf{a}\}| + |\{\sigma'\}|}{2}
\end{equation}

Thus, side networks establish a new efficiency frontier for Memory-Efficient Transfer Learning (METL), with memory savings growing linearly with $r$ while maintaining functional capacity.

\section{Baselines}
\label{sec:supple_baselines}
We select various transfer paradigms for comprehensive and challenging validation:

-\textit{VSE$\infty$} \cite{chen2021VSE} with BERT-base \cite{devllin2019bert} model and ResNeXt-101(32$\times$8d) \cite{xie2017aggregated} backbone pre-trained on Instagram (WSL) on Flickr30K \cite{young2014image}, MSCOCO1K and MSCOCO5K \cite{lin2014microsoft} for the \textbf{ITR} task;

-\textit{CLIP4Clip} \cite{luo2021clip4clip} with the pre-trained CLIP \cite{radford2021learning} using Text Transformer \cite{radford2019language} and ViT-B/32 \cite{alexey2021vit} on MSR-VTT \cite{xu2016msr} and MSVD \cite{chen2011collecting} for the \textbf{VTR} task;

-\textit{CLIP-ViL} \cite{shen2021much} that applies the CLIP image backbone \cite{radford2021learning} and encodes the text into word embeddings, followed by a cross-modal Transformer on VQAv2 \cite{goyal2017making} and GQA \cite{hudson2019gqa} for the \textbf{QA} task;

-\textit{MDETR} \cite{kamath2021mdetr} that integrates a pre-trained ResNet-101, RoBERTa-base \cite{liu2019roberta} with an encoder-decoder Transformer on RefCOCO, RefCOCO+ \cite{yu2016modeling} and RefCOCOg \cite{mao2016generation} for the \textbf{VG} task;

-\textit{T5-series} \cite{raffel2020exploring} that imports text encoder and auto-regressive decoder, while following \cite{sung2022lst}, we drop 6, 24 layers of side network (3, 12 layers each in encoder and decoder) for \textit{T5-base} and \textit{T5-large} on GLUE benchmark \cite{wang2018glue} for the \textbf{NLP} task;

\textit{ViT-base} \cite{alexey2021vit} witch consists of 86 million parameters, while pre-trained on ImageNet-21K \cite{deng2009imagenet} is the most commonly used backbone across prior works (\emph{e.g.}, image classification, video classification, \emph{etc}.), and is adopted on VTAB-1K \cite{zhai2019large} for the \textbf{CV} task.

\section{More Ablation Studies}
\label{sec:supple_ablation}
We extensively conduct more ablation studies to verify the effectiveness of our proposed method and the selected hyper-parameters.

\begin{figure*}[!t]
    \centering
    \includegraphics[width=1\textwidth]{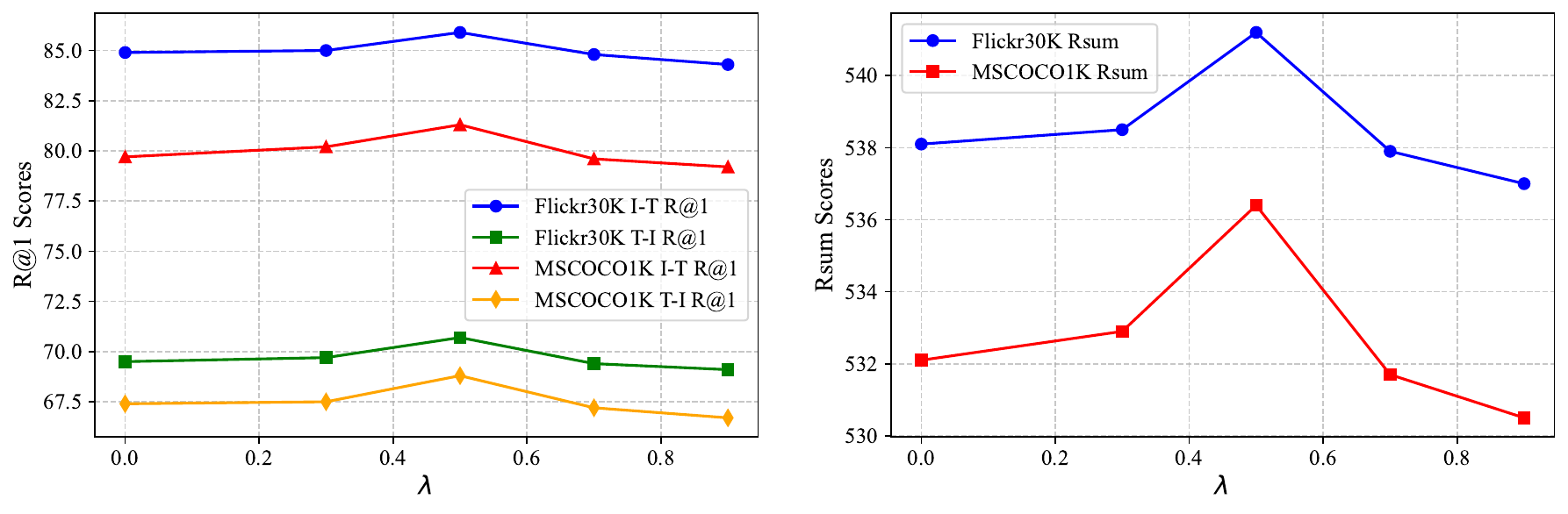} 
    \caption{Ablation study on the effect of the mask rate $\lambda$ in our method. ($\textbf{Left}$) The Rsum (\%) performance on Flickr30K and MSCOCO1K datasets. ($\textbf{Right}$) The R@1 (\%) performance for sentence retrieval ("I-T") and image retrieval ("T-I") on Flickr30K and MSCOCO1K datasets.}
    \label{fig_lambda_ablation}
\end{figure*}

\begin{table*}[!t]
    \centering
    \caption{Ablation results (\%) on the generation blocks for feature transformation in the student network on distinct datasets for ITR task. The best results are highlighted in \textbf{bold}.} 
    { \begin{tabular}{ccc ccc ccc ccc}
         \toprule[1.2pt]
         \multirow{2}{*}{\makecell{Generation\\Blocks}} & {Params.} & {Memory} & \multicolumn{3}{c}{Flickr30K} & \multicolumn{3}{c}{MSCOCO1K} & \multicolumn{3}{c}{MSCOCO5K} \\
         \cmidrule[0.4pt]{4-12}
         & (M) $\downarrow$ & (G) $\downarrow$ & {I-T} $\uparrow$ & {T-I} $\uparrow$ & {Rsum} $\uparrow$ & {I-T} $\uparrow$ & {T-I} $\uparrow$ & {Rsum} $\uparrow$ & {I-T} $\uparrow$ & {T-I} $\uparrow$ & {Rsum} $\uparrow$ \\
         \midrule[0.6pt]
         Self-Attention & 12.6 & 25.4 & 84.7 & 69.5 & 537.4 & 79.8 & 67.2 & 531.5 & 60.4 & 46.4 & 444.9 \\
         {Convolution} & \textbf{12.4} & \textbf{25.1} & \textbf{85.9} & \textbf{70.7} & \textbf{541.2} & \textbf{81.3} & \textbf{68.8} & \textbf{536.4} & \textbf{61.9} & \textbf{46.8} & \textbf{448.1} \\
         \bottomrule[1.2pt]
    \end{tabular}
    }
    \label{tab_ablation_gen_blocks}
\end{table*}

\begin{table*}[!t]
    \centering
    \caption{Ablation results (\%) on adopting Hierarchical Feature-based Distillation strategy across shallow \textbf{(Top)}, deep \textbf{(Middle)}, and all \textbf{(Bottom)} layers on distinct datasets for ITR task. The best results are highlighted in \textbf{bold}. and the second best results are \underline{underlined}.}
    {\begin{tabular}{cccc ccc ccc ccc}
         \toprule[1.2pt]
         \multicolumn{2}{c}{\multirow{2}{*}{Layers}} & {Params.} & {Memory} & \multicolumn{3}{c}{Flickr30K} & \multicolumn{3}{c}{MSCOCO1K} & \multicolumn{3}{c}{MSCOCO5K} \\
         \cmidrule[0.4pt]{5-13}
         & & (M) $\downarrow$ & (G) $\downarrow$ & {I-T} $\uparrow$ & {T-I} $\uparrow$ & {Rsum} $\uparrow$ & {I-T} $\uparrow$ & {T-I} $\uparrow$ & {Rsum} $\uparrow$ & {I-T} $\uparrow$ & {T-I} $\uparrow$ & {Rsum} $\uparrow$ \\
         \midrule[0.6pt]
         \multirow{5}{*}{{\rotatebox {90}{Imitation}}} & 1 & \textbf{12.4} & \textbf{24.9} & 83.4 & 68.5 & 536.8 & 78.6 & 65.8 & 529.1 & 59.1 & 44.7 & 441.3 \\
         & 1,2 & \textbf{12.4} & \textbf{24.9} & 83.7 & 68.8 & 536.7 & 78.9 & 66.1 & 529.2 & 59.7 & 44.3 & 441.6 \\
         & 5,6 & \textbf{12.4} & \textbf{24.9} & 84.6 & 69.2 & 537.4 & 79.6 & 67.1 & 531.2 & 60.5 & \underline{45.6} & 443.8 \\
         & {1 $\sim$ 6} & \textbf{12.4} & {25.1} & 84.5 & 69.5 & 537.6 & 79.4 & 67.3 & 530.9 & 60.8 & 45.5 & 444.0 \\
         & \textcolor{gray}{All} & \textcolor{gray}{12.4} & \textcolor{gray}{25.1} & \textcolor{gray}{84.9} & \textcolor{gray}{69.7} & \textcolor{gray}{538.3} & \textcolor{gray}{80.0} & \textcolor{gray}{67.8} & \textcolor{gray}{532.6} & \textcolor{gray}{61.4} & \textcolor{gray}{45.8} & \textcolor{gray}{445.7} \\
         \midrule[0.6pt]
         \multirow{5}{*}{{\rotatebox {90}{Generation}}} & 7,8 & \textbf{12.4} & \underline{25.0} & 84.6 & 69.2 & 537.6 & 79.3 & 66.5 & 530.2 & 60.8 & 44.5 & 442.9 \\
         & 11,12 & \textbf{12.4} & \underline{25.0} & 85.0 & 69.5 & 538.4 & 79.8 & 67.1 & 531.3 & \underline{61.4} & 44.8 & 444.3 \\
         & 12 & \textbf{12.4} & \underline{25.0} & \underline{85.2} & \underline{69.8} & \underline{538.6} & \underline{80.2} & \underline{67.4} & \underline{532.6} & 61.2 & 45.4 & \underline{445.2} \\
         & {7 $\sim$ 12} & \textbf{12.4} & {25.2} & 84.6 & 69.3 & 538.1 & 79.9 & 67.2 & 531.6 & 60.7 & 45.2 & 444.1 \\
         & \textcolor{gray}{All} & \textcolor{gray}{12.4} & \textcolor{gray}{25.4} & \textcolor{gray}{85.2} & \textcolor{gray}{70.1} & \textcolor{gray}{539.5} & \textcolor{gray}{80.6} & \textcolor{gray}{67.9} & \textcolor{gray}{533.4} & \textcolor{gray}{61.1} & \textcolor{gray}{45.9} & \textcolor{gray}{446.0} \\
         \midrule[0.6pt]
         \multicolumn{2}{c}{All} & \textbf{12.4} & {25.1} & \textbf{85.9} & \textbf{70.7} & \textbf{541.2} & \textbf{81.3} & \textbf{68.8} & \textbf{536.4} & \textbf{61.9} & \textbf{46.8} & \textbf{448.1} \\
         \bottomrule[1.2pt]
    \end{tabular}}
    \label{tab_ablation_layer_norm}
\end{table*}

\paragraph{Influence of the mask ratio $\bm{\lambda}$.}
To explore the impact of the mask ratio and verify the effectiveness of adopting $\lambda = 0.5$, we conduct an ablation study of $\lambda$ on ITR task. As demonstrated in Figure~\ref{fig_lambda_ablation}, the results reveal that the performance of our method is sensitive to $\lambda$. Specifically, the model achieves the highest accuracy when $\lambda = 0.5$ on both Flickr30K and MSCOCO1K datasets, indicating that $\lambda = 0.5$ optimally balances the preservation of critical information and the introduction of diversity during feature distillation. As $\lambda$ increases beyond 0.5, the performance of our method gradually declines. This degradation can be attributed to the excessive masking, which leads to the loss of valuable information, thereby diminishing the effectiveness of feature distillation and impairing the student network's ability to learn useful information from the teacher network. 

\paragraph{Necessity of generation blocks.}
We adopt a convolutional projector as the generation module for feature transformation in the student network. In order to validate its effectiveness, we conduct an ablation study comparing it with the self-attention mechanism. As shown in Table~\ref{tab_ablation_gen_blocks}, the convolutional projector achieves superior performance on ITR task, demonstrating its effectiveness in mitigating the feature discrepancies between the teacher and student networks. Beyond its advantage in accuracy, the convolutional projector also incurs remarkable advantages in computational efficiency, as it requires fewer parameters and consumes less training memory. These benefits stem from their fundamental differences in computational design. The self-attention mechanism computes attention scores for all input element pairs, leading to quadratic complexity with respect to the input size. This design significantly increases memory consumption and computational overhead during training. In contrast, the convolutional module operates on local receptive fields, capturing spatially localized features with a linear computational complexity with respect to the input size. This efficiency makes the convolutional projector computationally lightweight, thus being more suitable for applications on resource-constrained scenarios for large-scale tasks.

\paragraph{Necessity of distillation across all layers.}
To verify the necessity of performing feature distillation across all layers, we empirically investigate the performance of our method with different layer combinations when performing distillation. As shown in Table~\ref{tab_ablation_layer_norm}, the results indicate that performing Hierarchical Feature-based Distillation strategy across all layers yields the highest accuracy on ITR task. Furthermore, when feature distillation is applied solely to the shallow layers (\emph{i.e.}, layer 1$\sim$6), the performance is notably lower compared to distillation applied to the deep layers (\emph{i.e.}, layer 7$\sim$12). The performance degradation is attributed to the fact that shallow layers contain less semantic information, which limits their contribution to the overall task. In contrast, deeper layers encode semantically rich features, leading to superior results. {To further demonstrate the effectiveness of HFD independently, we also conduct ablation experiments with \textit{imitation-only} and \textit{generation-only} (\emph{i.e.,} feature-based distillation is performed across all layers utilizing either imitation or generation methods), the results are indicated in gray.} These results underscore the importance of leveraging features from all layers of the network during the distillation process, providing evidence that our approach effectively facilitates the knowledge transfer from the backbone to the side network.

%% file: CVPR26_MDPD/main.bib
@String(NIPS= {Adv. Neural Inform. Process. Syst.})

@String(AAAI = {AAAI})

@String(NIPS  = {NeurIPS})

@inproceedings{alexey2021vit,
   author = {Dosovitskiy, Alexey and Beyer, Lucas and Kolesnikov, Alexander and Weissenborn, Dirk and Zhai, Xiaohua and Unterthiner, Thomas and Dehghani, Mostafa and Minderer, Matthias and Heigold, Georg and Gelly, Sylvain and Uszkoreit, Jakob and Houlsby, Neil},
   title = {An Image is Worth 16x16 Words: Transformers for Image Recognition at Scale},
   booktitle = {Proceedings of the International Conference on Learning Representations},
   year = {2021},
   type = {Conference Proceedings}
}

@inproceedings{fang2023eva,
   author = {Fang, Yuxin and Wang, Wen and Xie, Binhui and Sun, Quan and Wu, Ledell and Wang, Xinggang and Huang, Tiejun and Wang, Xinlong and Cao, Yue},
   title = {EVA: Exploring the Limits of Masked Visual Representation Learning at Scale},
   booktitle = {Proceedings of the Computer Vision and Pattern Recognition},
   pages = {19358-19369},
   year = {2023},
   type = {Conference Proceedings}
}

@article{liu2019roberta,
   author = {Liu, Yinhan and Ott, Myle and Goyal, Naman and Du, Jingfei and Joshi, Mandar and Chen, Danqi and Levy, Omer and Lewis, Mike and Zettlemoyer, Luke and Stoyanov, Veselin},
   title = {RoBERTa: A Robustly Optimized BERT Pretraining Approach},
   journal = {arXiv preprint arXiv:.11692},
   volume = {364},
   year = {2019},
   type = {Journal Article}
}

@inproceedings{devllin2019bert,
   author = {Devlin, Jacob and Chang, Ming-Wei and Lee, Kenton and Toutanova, Kristina},
   title = {BERT: Pre-training of Deep Bidirectional Transformers for Language Understanding},
   booktitle = {Proceedings of the North American Chapter of the Association for Computational Linguistics},
   pages = {4171-4186},
   year = {2019},
   type = {Conference Proceedings}
}

@article{raffel2020exploring,
   author = {Raffel, Colin and Shazeer, Noam and Roberts, Adam and Lee, Katherine and Narang, Sharan and Matena, Michael and Zhou, Yanqi and Li, Wei and Liu, Peter J},
   title = {Exploring the Limits of Transfer Learning with a Unified Text-to-Text Transformer},
   journal = {Journal of Machine Learning Research},
   volume = {21},
   number = {140},
   pages = {1-67},
   ISSN = {1533-7928},
   year = {2020},
   type = {Journal Article}
}

@inproceedings{girdhar2023image,
   author = {Girdhar, Rohit and El-Nouby, Alaaeldin and Liu, Zhuang and Singh, Mannat and Alwala, Kalyan Vasudev and Joulin, Armand and Misra, Ishan},
   title = {ImageBind One Embedding Space to Bind Them All},
   booktitle = {Proceedings of the Computer Vision and Pattern Recognition},
   pages = {15180-15190},
   ISBN = {2575-7075},
   year = {2023},
   type = {Conference Proceedings}
}

@inproceedings{radford2021learning,
   author = {Radford, Alec and Kim, Jong Wook and Hallacy, Chris and Ramesh, Aditya and Goh, Gabriel and Agarwal, Sandhini and Sastry, Girish and Askell, Amanda and Mishkin, Pamela and Clark, Jack},
   title = {Learning Transferable Visual Models from Natural Language Supervision},
   booktitle = {Proceedings of the International Conference on Machine Learning},
   pages = {8748-8763},
   ISBN = {2640-3498},
   year = {2021},
   type = {Conference Proceedings}
}

@inproceedings{cai2020tinytl,
   author = {Cai, Han and Gan, Chuang and Zhu, Ligeng and Han, Song},
   title = {Tinytl: Reduce Memory, Not Parameters for Efficient On-Device Learning},
   booktitle = {Proceedings of the Advances in Neural Information Processing Systems},
   pages = {11285-11297},
   year = {2020},
   type = {Conference Proceedings}
}

@inproceedings{zaken2022bitfit,
   author = {Zaken, Elad Ben and Ravfogel, Shauli and Goldberg, Yoav},
   title = {BitFit: Simple Parameter-Efficient Fine-Tuning for Transformer-Based Masked Language-Models},
   booktitle = {Proceedings of the Annual Meeting of the Association for Computational Linguistics},
   pages = {1--9},
   year = {2022},
   type = {Conference Proceedings}
}

@inproceedings{kim2021howto,
   author = {Kim, Konwoo and Laskin, Michael and Mordatch, Igor and Pathak, Deepak},
   title = {How to Adapt Your Large-Scale Vision-and-Language Model},
   booktitle = {Proceedings of the International Conference on Learning Representations},
   year = {2021},
   type = {Conference Proceedings}
}

@inproceedings{touvron2022three,
   author = {Touvron, Hugo and Cord, Matthieu and El-Nouby, Alaaeldin and Verbeek, Jakob and Jégou, Hervé},
   title = {Three Things Everyone Should Know about Vision Transformers},
   booktitle = {Proceedings of the European Conference on Computer Vision},
   pages = {497-515},
   year = {2022},
   type = {Conference Proceedings}
}

@inproceedings{zhao2020masking,
   author = {Zhao, Mengjie and Lin, Tao and Mi, Fei and Jaggi, Martin and Schütze, Hinrich},
   title = {Masking as an Efficient Alternative to Finetuning for Pretrained Language Models},
   booktitle = {Proceedings of the Conference on Empirical Methods in Natural Language Processing},
   year = {2020},
   type = {Conference Proceedings}
}

@inproceedings{houlsby2019param,
   author = {Houlsby, Neil and Giurgiu, Andrei and Jastrzebski, Stanislaw and Morrone, Bruna and Laroussilhe, Quentin De and Gesmundo, Andrea and Attariyan, Mona and Gelly, Sylvain},
   title = {Parameter-Efficient Transfer Learning for NLP},
   booktitle = {Proceedings of the International Conference on Machine Learning},
   pages = {2790--2799},
   year = {2019},
   type = {Conference Proceedings}
}

@inproceedings{he2022towards,
   author = {He, Junxian and Zhou, Chunting and Ma, Xuezhe and Berg-Kirkpatrick, Taylor and Neubig, Graham},
   title = {Towards a Unified View of Parameter-Efficient Transfer Learning},
   booktitle = {Proceedings of the International Conference on Learning Representations},
   year = {2022},
   type = {Conference Proceedings}
}

@inproceedings{zhu2021counter,
   author = {Zhu, Yaoming and Feng, Jiangtao and Zhao, Chengqi and Wang, Mingxuan and Li, Lei},
   title = {Counter-Interference Adapter for Multilingual Machine Translation},
   booktitle = {Proceedings of the Conference on Empirical Methodsin Natural Language Processing},
   pages = {2812-2823},
   year = {2021},
   type = {Conference Proceedings}
}

@inproceedings{he2022sparse,
   author = {He, Shwai and Ding, Liang and Dong, Daize and Zhang, Miao and Tao, Dacheng},
   title = {SparseAdapter: An Easy Approach for Improving the Parameter-Efficiency of Adapters},
   booktitle = {Proceedings of the Conference on Empirical Methodsin Natural Language Processing},
   pages = {2184-2190},
   year = {2022},
   type = {Conference Proceedings}
}

@inproceedings{he2023sensitivity,
   author = {He, Haoyu and Cai, Jianfei and Zhang, Jing and Tao, Dacheng and Zhuang, Bohan},
   title = {Sensitivity-Aware Visual Parameter-Efficient Fine-Tuning},
   booktitle = {Proceedings of the IEEE International Conference on Computer Vision},
   pages = {11825-11835},
   year = {2023},
   type = {Conference Proceedings}
}

@inproceedings{lester2021power,
   author = {Lester, Brian and Al-Rfou, Rami and Constant, Noah},
   title = {The Power of Scale for Parameter-Efficient Prompt Tuning},
   booktitle = {Proceedings of the Conference on Empirical Methods in Natural Language Processing},
   pages = {3045-3059},
   year = {2021},
   type = {Conference Proceedings}
}

@inproceedings{li2021prefix,
   author = {Li, Xiang Lisa and Liang, Percy},
   title = {Prefix-Tuning: Optimizing Continuous Prompts for Generation},
   booktitle = {Proceedings of the Annual Meeting of the Association for Computational Linguistics},
   pages = {4582--4597},
   year = {2021},
   type = {Conference Proceedings}
}

@inproceedings{vu2021spot,
   author = {Vu, Tu and Lester, Brian and Constant, Noah and Al-Rfou, Rami and Cer, Daniel},
   title = {SPoT: Better Frozen Model Adaptation Through Soft Prompt Transfer},
   booktitle = {Proceedings of the Annual Meeting of the Association for Computational Linguistics},
   year = {2021},
   type = {Conference Proceedings}
}

@inproceedings{hambardzumyan2021warp,
   author = {Hambardzumyan, Karen and Khachatrian, Hrant and May, Jonathan},
   title = {WARP: Word-Level Adversarial Reprogramming},
   booktitle = {Proceedings of the Annual Meeting of the Association for Computational Linguistics},
   year = {2021},
   type = {Conference Proceedings}
}

@inproceedings{qin2022exploring,
   author = {Qin, Yujia and Wang, Xiaozhi and Su, Yusheng and Lin, Yankai and Ding, Ning and Yi, Jing and Chen, Weize and Liu, Zhiyuan and Li, Juanzi and Hou, Lei},
   title = {Exploring Universal Intrinsic Task Subspace via Prompt Tuning},
   booktitle = {Proceedings of the Annual Meeting of the Association for Computational Linguistics},
   year = {2022},
   type = {Conference Proceedings}
}

@inproceedings{hu2022lora,
   author = {Hu, Edward J and Shen, Yelong and Wallis, Phillip and Allen-Zhu, Zeyuan and Li, Yuanzhi and Wang, Shean and Wang, Lu and Chen, Weizhu},
   title = {LoRA: Low-Rank Adaptation of Large Language Models},
   booktitle = {Proceedings of the International Conference on Learning Representations},
   year = {2022},
   type = {Conference Proceedings}
}

@inproceedings{he2023param,
   author = {He, Xuehai and Li, Chunyuan and Zhang, Pengchuan and Yang, Jianwei and Wang, Xin Eric},
   title = {Parameter-Efficient Model Adaptation for Vision Transformers},
   booktitle = {Proceedings of the AAAI Conference on Artificial Intelligence},
   pages = {817-825},
   year = {2023},
   type = {Conference Proceedings}
}

@inproceedings{song2024increasing,
   author = {Song, Haobo and Zhao, Hao and Majumder, Soumajit and Lin, Tao},
   title = {Increasing Model Capacity for Free: A Simple Strategy for Parameter Efficient Fine-Tuning},
   booktitle = {Proceedings of the International Conference on Learning Representations},
   year = {2024},
   type = {Conference Proceedings}
}

@article{wu2024mixture,
   author = {Wu, Xun and Huang, Shaohan and Wei, Furu},
   title = {Mixture of LoRA Experts},
   journal = {arXiv preprint arXiv:.13628},
   year = {2024},
   type = {Journal Article}
}

@inproceedings{zhang2020side,
   author = {Zhang, Jeffrey O and Sax, Alexander and Zamir, Amir and Guibas, Leonidas and Malik, Jitendra},
   title = {Side-Tuning: A Baseline for Network Adaptation via Additive Side Networks},
   booktitle = {Proceedings of the European Conference on Computer Vision},
   pages = {698-714},
   year = {2020},
   type = {Conference Proceedings}
}

@inproceedings{sung2022lst,
   author = {Sung, Yi-Lin and Cho, Jaemin and Bansal, Mohit},
   title = {LST: Ladder Side-Tuning for Parameter and Memory Efficient Transfer Learning},
   booktitle = {Proceedings of the Advances in Neural Information Processing Systems},
   pages = {12991-13005},
   year = {2022},
   type = {Conference Proceedings}
}

@inproceedings{liao2023make,
   author = {Liao, Baohao and Tan, Shaomu and Monz, Christof},
   title = {Make Pre-trained Model Reversible: From Parameter to Memory Efficient Fine-Tuning},
   booktitle = {Proceedings of the Advances in Neural Information Processing Systems},
   year = {2023},
   type = {Conference Proceedings}
}

@inproceedings{gomez2017reversible,
   author = {Gomez, Aidan N and Ren, Mengye and Urtasun, Raquel and Grosse, Roger B },
   title = {The Reversible Residual Network: Backpropagation Without Storing Activations},
   booktitle = {Proceedings of the Advances in Neural Information Processing Systems},
   year = {2017},
   type = {Conference Proceedings}
}

@article{zhang2023lora-fa,
   author = {Zhang, Longteng and Zhang, Lin and Shi, Shaohuai and Chu, Xiaowen and Li, Bo},
   title = {LoRA-FA: Memory-Efficient Low-rank Adaptation for Large Language Models Fine-tuning},
   journal = {arXiv preprint arXiv:.03303},
   year = {2023},
   type = {Journal Article}
}

@inproceedings{phang2023hypertuning,
   author = {Phang, Jason and Mao, Yi and He, Pengcheng and Chen, Weizhu},
   title = {HyperTuning: Toward Adapting Large Language Models without Back-propagation},
   booktitle = {Proceedings of the International Conference on Machine Learning},
   pages = {27854-27875},
   year = {2023},
   type = {Conference Proceedings}
}

@inproceedings{malladi2023tune,
   author = {Malladi, Sadhika and Gao, Tianyu and Nichani, Eshaan and Damian, Alex and Lee, Jason D and Chen, Danqi and Arora, Sanjeev},
   title = {Fine-Tuning Language Models with Just Forward Passes},
   booktitle = {Proceedings of the Advances in Neural Information Processing Systems},
   pages = {53038-53075},
   year = {2023},
   type = {Conference Proceedings}
}

@article{hinton2015distilling,
   author = {Hinton, Geoffrey},
   title = {Distilling the Knowledge in a Neural Network},
   journal = {arXiv preprint arXiv:.02531},
   year = {2015},
   type = {Journal Article}
}

@inproceedings{romero2015fitnets,
   author = {Romero, Adriana and Ballas, Nicolas and Kahou, Samira Ebrahimi and Chassang, Antoine and Gatta, Carlo and Bengio, Yoshua },
   title = {FitNets: Hints for Thin Deep Nets},
   booktitle = {Proceedings of the International Conference on Learning Representations},
   year = {2015},
   type = {Conference Proceedings}
}

@inproceedings{chen2021cross,
   author = {Chen, Defang and Mei, Jian-Ping and Zhang, Yuan and Wang, Can and Wang, Zhe and Feng, Yan and Chen, Chun},
   title = {Cross-Layer Distillation with Semantic Calibration},
   booktitle = {Proceedings of the AAAI Conference on Artificial Intelligence},
   pages = {7028-7036},
   year = {2021},
   type = {Conference Proceedings}
}

@inproceedings{yim2017gift,
   author = {Yim, Junho and Joo, Donggyu and Bae, Jihoon and Kim, Junmo},
   title = {A Gift from Knowledge Distillation: Fast Optimization, Network Minimization and Transfer Learning},
   booktitle = {Proceedings of the IEEE Conference on Computer Vision and Pattern Recognition},
   pages = {4133-4141},
   year = {2017},
   type = {Conference Proceedings}
}

@article{chen2021learning,
   author = {Chen, Hanting and Wang, Yunhe and Xu, Chang and Xu, Chao and Tao, Dacheng},
   title = {Learning Student Networks via Feature Embedding},
   journal = {IEEE Transactions on Neural Networks Learning Systems},
   volume = {32},
   number = {1},
   pages = {25-35},
   ISSN = {2162-237X},
   year = {2021},
   type = {Journal Article}
}

@inproceedings{meng2019conditional,
   author = {Meng, Zhong and Li, Jinyu and Zhao, Yong and Gong, Yifan},
   title = {Conditional Teacher-Student Learning},
   booktitle = {Proceedings of the IEEE International Conference on Acoustics, Speech and Signal Processing},
   pages = {6445-6449},
   year = {2019},
   type = {Conference Proceedings}
}

@inproceedings{wu2022tinyvit,
   author = {Wu, Kan and Zhang, Jinnian and Peng, Houwen and Liu, Mengchen and Xiao, Bin and Fu, Jianlong and Yuan, Lu},
   title = {TinyViT: Fast Pretraining Distillation for Small Vision Transformers},
   booktitle = {Proceedings of the European Conference on Computer Vision},
   pages = {68-85},
   year = {2022},
   type = {Conference Proceedings}
}

@article{diao2024unveiling,
  title={Unveiling Encoder-Free Vision-Language Models},
  author={Diao, Haiwen and Cui, Yufeng and Li, Xiaotong and Wang, Yueze and Lu, Huchuan and Wang, Xinlong},
  journal={arXiv preprint arXiv:2406.11832},
  year={2024}
}

@inproceedings{li2023blip,
  title={Blip-2: Bootstrapping language-image pre-training with frozen image encoders and large language models},
  author={Li, Junnan and Li, Dongxu and Savarese, Silvio and Hoi, Steven},
  booktitle={Proceedings of the International Conference on Machine Learning},
  pages={19730--19742},
  year={2023}
}

@article{guo2020parameter,
  title={Parameter-efficient transfer learning with diff pruning},
  author={Guo, Demi and Rush, Alexander M and Kim, Yoon},
  journal={arXiv preprint arXiv:2012.07463},
  year={2020}
}

@inproceedings{jia2022visual,
  title={Visual prompt tuning},
  author={Jia, Menglin and Tang, Luming and Chen, Bor-Chun and Cardie, Claire and Belongie, Serge and Hariharan, Bharath and Lim, Ser-Nam},
  booktitle={Proceedings of the European Conference on Computer Vision},
  pages={709--727},
  year={2022}
}

@article{liu2024tuning,
  title={Y-tuning: An efficient tuning paradigm for large-scale pre-trained models via label representation learning},
  author={Liu, Yitao and An, Chenxin and Qiu, Xipeng},
  journal={Frontiers of Computer Science},
  volume={18},
  number={4},
  pages={184320},
  year={2024},
  publisher={Springer}
}

@article{chen2022vision,
  title={Vision transformer adapter for dense predictions},
  author={Chen, Zhe and Duan, Yuchen and Wang, Wenhai and He, Junjun and Lu, Tong and Dai, Jifeng and Qiao, Yu},
  journal={arXiv preprint arXiv:2205.08534},
  year={2022}
}

@inproceedings{chen2022adaptformer,
  title={Adaptformer: Adapting vision transformers for scalable visual recognition},
  author={Chen, Shoufa and Ge, Chongjian and Tong, Zhan and Wang, Jiangliu and Song, Yibing and Wang, Jue and Luo, Ping},
  booktitle={Proceedings of the Advances in Neural Information Processing Systems},
  pages={16664--16678},
  year={2022}
}

@article{zhou2022learning,
  title={Learning to prompt for vision-language models},
  author={Zhou, Kaiyang and Yang, Jingkang and Loy, Chen Change and Liu, Ziwei},
  journal={International Journal of Computer Vision},
  volume={130},
  number={9},
  pages={2337--2348},
  year={2022},
  publisher={Springer}
}

@inproceedings{sung2021training,
  title={Training neural networks with fixed sparse masks},
  author={Sung, Yi-Lin and Nair, Varun and Raffel, Colin A},
  booktitle={Proceedings of the Advances in Neural Information Processing Systems},
  pages={24193--24205},
  type = {Conference Proceedings},
  year={2021}
}

@article{sun2023generative,
  title={Generative pretraining in multimodality},
  author={Sun, Quan and Yu, Qiying and Cui, Yufeng and Zhang, Fan and Zhang, Xiaosong and Wang, Yueze and Gao, Hongcheng and Liu, Jingjing and Huang, Tiejun and Wang, Xinlong},
  journal={arXiv preprint arXiv:2307.05222},
  year={2023}
}

@inproceedings{he2022masked,
  title={Masked autoencoders are scalable vision learners},
  author={He, Kaiming and Chen, Xinlei and Xie, Saining and Li, Yanghao and Doll{\'a}r, Piotr and Girshick, Ross},
  booktitle={Proceedings of the Computer Vision and Pattern Recognition},
  pages={16000--16009},
  year={2022}
}

@article{touvron2023llama,
  title={Llama: Open and efficient foundation language models},
  author={Touvron, Hugo and Lavril, Thibaut and Izacard, Gautier and Martinet, Xavier and Lachaux, Marie-Anne and Lacroix, Timoth{\'e}e and Rozi{\`e}re, Baptiste and Goyal, Naman and Hambro, Eric and Azhar, Faisal and others},
  journal={arXiv preprint arXiv:2302.13971},
  year={2023}
}

@inproceedings{bai2023masked,
   author = {Bai, Yutong and Wang, Zeyu and Xiao, Junfei and Wei, Chen and Wang, Huiyun and Yuille, Alan and Zhou, Yuyin and Xie, Cihang},
   title = {Masked Autoencoders Enable Efficient Knowledge Distillers},
   booktitle = {Proceedings of the IEEE Conference on Computer Vision and Pattern Recognition},
   pages = {24256-24265},
   year = {2023},
   type = {Conference Proceedings}
}

@inproceedings{diao2024unipt,
   author = {Diao, Haiwen and Wan, Bo and Zhang, Ying and Jia, Xu and Lu, Huchuan and Chen, Long},
   title = {UniPT: Universal Parallel Tuning for Transfer Learning with Efficient Parameter and Memory},
   booktitle = {Proceedings of the Computer Vision and Pattern Recognition},
   pages = {28729-28740},
   year = {2024},
   type = {Conference Proceedings}
}

@inproceedings{diao2024sherl,
   author = {Diao, Haiwen and Wan, Bo and Jia, Xu and Zhuge, Yunzhi and Zhang, Ying and Lu, Huchuan and Chen, Long},
   title = {SHERL: Synthesizing High Accuracy and Efficient Memory for Resource-Limited Transfer Learning},
   booktitle = {Proceedings of the European Conference on Computer Vision},
   pages = {75-95},
   year = {2024},
   type = {Conference Proceedings}
}

@article{zhai2019large,
  title={A large-scale study of representation learning with the visual task adaptation benchmark},
  author={Zhai, Xiaohua and Puigcerver, Joan and Kolesnikov, Alexander and Ruyssen, Pierre and Riquelme, Carlos and Lucic, Mario and Djolonga, Josip and Pinto, Andre Susano and Neumann, Maxim and Dosovitskiy, Alexey and others},
  journal={arXiv preprint arXiv:1910.04867},
  year={2019},
  type = {Journal Article}
}

@inproceedings{lin2014microsoft,
  title={Microsoft coco: Common objects in context},
  author={Lin, Tsung-Yi and Maire, Michael and Belongie, Serge and Hays, James and Perona, Pietro and Ramanan, Deva and Doll{\'a}r, Piotr and Zitnick, C Lawrence},
  booktitle={Proceedings of the European Conference on Computer Vision},
  pages={740--755},
  year={2014},
  type = {Conference Proceedings}
}

@article{young2014image,
  title={From image descriptions to visual denotations: New similarity metrics for semantic inference over event descriptions},
  author={Young, Peter and Lai, Alice and Hodosh, Micah and Hockenmaier, Julia},
  journal={Transactions of the Association for Computational Linguistics},
  volume={2},
  pages={67--78},
  year={2014},
  type = {Journal Article}
}

@inproceedings{chen2011collecting,
  title={Collecting highly parallel data for paraphrase evaluation},
  author={Chen, David and Dolan, William B},
  booktitle={Proceedings of the Annual Meeting of the Association for Computational Linguistics},
  pages={190--200},
  year={2011},
  type = {Conference Proceedings}
}

@inproceedings{xu2016msr,
  title={Msr-vtt: A large video description dataset for bridging video and language},
  author={Xu, Jun and Mei, Tao and Yao, Ting and Rui, Yong},
  booktitle={Proceedings of the IEEE Conference on Computer Vision and Pattern Recognition},
  pages={5288--5296},
  year={2016},
  type = {Conference Proceedings}
}

@inproceedings{yu2016modeling,
  title={Modeling context in referring expressions},
  author={Yu, Licheng and Poirson, Patrick and Yang, Shan and Berg, Alexander C and Berg, Tamara L},
  booktitle={Proceedings of the European Conference on Computer Vision},
  pages={69--85},
  year={2016},
  type = {Conference Proceedings}
}

@inproceedings{mao2016generation,
  title={Generation and comprehension of unambiguous object descriptions},
  author={Mao, Junhua and Huang, Jonathan and Toshev, Alexander and Camburu, Oana and Yuille, Alan L and Murphy, Kevin},
  booktitle={Proceedings of the Computer Vision and Pattern Recognition},
  pages={11--20},
  year={2016}
}

@inproceedings{chen2021VSE,
  title={Learning the best pooling strategy for visual semantic embedding},
  author={Chen, Jiacheng and Hu, Hexiang and Wu, Hao and Jiang, Yuning and Wang, Changhu},
  booktitle={Proceedings of the IEEE Conference on Computer Vision and Pattern Recognition},
  pages={15789--15798},
  year={2021},
  type = {Conference Proceedings}
}

@article{luo2021clip4clip,
  title={Clip4clip: An empirical study of clip for end to end video clip retrieval},
  author={Luo, Huaishao and Ji, Lei and Zhong, Ming and Chen, Yang and Lei, Wen and Duan, Nan and Li, Tianrui},
  journal={arXiv preprint arXiv:2104.08860},
  year={2021},
  type = {Journal Article}
}

@inproceedings{xie2017aggregated,
  title={Aggregated residual transformations for deep neural networks},
  author={Xie, Saining and Girshick, Ross and Doll{\'a}r, Piotr and Tu, Zhuowen and He, Kaiming},
  booktitle={Proceedings of the Computer Vision and Pattern Recognition},
  pages={1492--1500},
  year={2017},
  type = {Conference Proceedings}
}

@article{radford2019language,
  title={Language models are unsupervised multitask learners},
  author={Radford, Alec and Wu, Jeffrey and Child, Rewon and Luan, David and Amodei, Dario and Sutskever, Ilya and others},
  journal={OpenAI blog},
  volume={1},
  number={8},
  pages={9},
  year={2019},
  type = {Journal Article}
}

@article{shen2021much,
  title={How much can clip benefit vision-and-language tasks?},
  author={Shen, Sheng and Li, Liunian Harold and Tan, Hao and Bansal, Mohit and Rohrbach, Anna and Chang, Kai-Wei and Yao, Zhewei and Keutzer, Kurt},
  journal={arXiv preprint arXiv:2107.06383},
  year={2021},
  type = {Journal Article}
}

@inproceedings{kamath2021mdetr,
  title={Mdetr-modulated detection for end-to-end multi-modal understanding},
  author={Kamath, Aishwarya and Singh, Mannat and LeCun, Yann and Synnaeve, Gabriel and Misra, Ishan and Carion, Nicolas},
  booktitle={Proceedings of the IEEE International Conference on Computer Vision},
  pages={1780--1790},
  year={2021}
}

@inproceedings{lian2022scaling,
  title={Scaling \& shifting your features: A new baseline for efficient model tuning},
  author={Lian, Dongze and Zhou, Daquan and Feng, Jiashi and Wang, Xinchao},
  booktitle={Proceedings of the Advances in Neural Information Processing Systems},
  pages={109--123}
}

@inproceedings{jie2023fact,
  title={Fact: Factor-tuning for lightweight adaptation on vision transformer},
  author={Jie, Shibo and Deng, Zhi-Hong},
  booktitle={Proceedings of the AAAI Conference on Artificial Intelligence},
  pages={1060--1068},
  year={2023}
}

@inproceedings{zhang2023adaptive,
  title={Adaptive Budget Allocation for Parameter-Efficient Fine-Tuning},
  author={Zhang, Qingru and Chen, Minshuo and Bukharin, Alexander and He, Pengcheng and Cheng, Yu and Chen, Weizhu and Zhao, Tuo},
  booktitle={Proceedings of the International Conference on Learning Representations},
  year={2023},
}

@inproceedings{mercea2024time,
  title={Time-Memory-and Parameter-Efficient Visual Adaptation},
  author={Mercea, Otniel-Bogdan and Gritsenko, Alexey and Schmid, Cordelia and Arnab, Anurag},
  booktitle={Proceedings of the Computer Vision and Pattern Recognition Conference},
  pages={5536--5545},
  year={2024}
}

@inproceedings{jiang2024res,
  title={Res-tuning: A flexible and efficient tuning paradigm via unbinding tuner from backbone},
  author={Jiang, Zeyinzi and Mao, Chaojie and Huang, Ziyuan and Ma, Ao and Lv, Yiliang and Shen, Yujun and Zhao, Deli and Zhou, Jingren},
  booktitle={Proceedings of the Advances in Neural Information Processing Systems},
  year={2024}
}

@article{lin2023hierarchical,
  title={Hierarchical side-tuning for vision transformers},
  author={Lin, Weifeng and Wu, Ziheng and Yang, Wentao and Huang, Mingxin and Huang, Jun and Jin, Lianwen},
  journal={arXiv preprint arXiv:2310.05393},
  year={2023}
}

@incollection{jie2024convolutional,
  title={Convolutional bypasses are better vision transformer adapters},
  author={Jie, Shibo and Deng, Zhi-Hong and Chen, Shixuan and Jin, Zhijuan},
  booktitle={Proceedings of the European Conference on Artificial Intelligence},
  pages={202--209},
  year={2024},
}

@article{2016Training,
  title={Training Deep Nets with Sublinear Memory Cost},
  author={Chen, Tianqi  and  Xu, Bing  and  Zhang, Chiyuan  and  Guestrin, Carlos },
  journal={arXiv preprint arXiv:1604.06174},
  year={2016},
}

@inproceedings{liang2024simple,
  title={Simple Yet Effective: Structure Guided Pre-trained Transformer for Multi-modal Knowledge Graph Reasoning},
  author={Liang, Ke and Meng, Lingyuan and Liu, Yue and Liu, Meng and Wei, Wei and Liu, Suyuan and Tu, Wenxuan and Wang, Siwei and Zhou, Sihang and Liu, Xinwang},
  booktitle={Proceedings of the ACM International Conference on Multimedia},
  pages={1554--1563},
  year={2024}
}

@inproceedings{zhuang2024towards,
  title={Towards Multimodal-augmented Pre-trained Language Models via Self-balanced Expectation-Maximization Iteration},
  author={Zhuang, Xianwei and Cheng, Xuxin and Zhu, Zhihong and Chen, Zhanpeng and Li, Hongxiang and Zou, Yuexian},
  booktitle={Proceedings of the ACM International Conference on Multimedia},
  pages={4670--4679},
  year={2024}
}

@inproceedings{liao2024uni,
  title={Uni-DlLoRA: Style Fine-Tuning for Fashion Image Translation},
  author={Liao, Fangjian and Zou, Xingxing and Wong, Waikeung},
  booktitle={Proceedings of the ACM International Conference on Multimedia},
  pages={6404--6413},
  year={2024}
}

@inproceedings{li2024multimodal,
  title={Multimodal Inplace Prompt Tuning for Open-set Object Detection},
  author={Li, Guilin and Zhang, Mengdan and Zheng, Xiawu and Chen, Peixian and Wang, Zihan and Shen, Yunhang and Zhuge, Mingchen and Wu, Chenglin and Chao, Fei and Li, Ke and others},
  booktitle={Proceedings of the ACM International Conference on Multimedia},
  pages={8062--8071},
  year={2024}
}

@inproceedings{yin2024parameter,
  title={Parameter-efficient is not sufficient: Exploring parameter, memory, and time efficient adapter tuning for dense predictions},
  author={Yin, Dongshuo and Han, Xueting and Li, Bin and Feng, Hao and Bai, Jing},
  booktitle={Proceedings of the ACM International Conference on Multimedia},
  pages={1398--1406},
  year={2024}
}

@inproceedings{pan2024disentangled,
  title={Disentangled-multimodal privileged knowledge distillation for depression recognition with incomplete multimodal data},
  author={Pan, Yuchen and Jiang, Junjun and Jiang, Kui and Liu, Xianming},
  booktitle={Proceedings of the ACM International Conference on Multimedia},
  pages={5712--5721},
  year={2024}
}

@inproceedings{chen2024joint,
  title={Joint Homophily and Heterophily Relational Knowledge Distillation for Efficient and Compact 3D Object Detection},
  author={Chen, Shidi and Wei, Lili and Liang, Liqian and Lang, Congyan},
  booktitle={Proceedings of the ACM International Conference on Multimedia},
  pages={2127--2135},
  year={2024}
}

@inproceedings{goyal2017making,
  title={Making the v in vqa matter: Elevating the role of image understanding in visual question answering},
  author={Goyal, Yash and Khot, Tejas and Summers-Stay, Douglas and Batra, Dhruv and Parikh, Devi},
  booktitle={Proceedings of the Computer Vision and Pattern Recognition Conference},
  pages={6904--6913},
  year={2017}
}

@inproceedings{hudson2019gqa,
  title={Gqa: A new dataset for real-world visual reasoning and compositional question answering},
  author={Hudson, Drew A and Manning, Christopher D},
  booktitle={Proceedings of the Computer Vision and Pattern Recognition Conference},
  pages={6700--6709},
  year={2019}
}

@article{wang2018glue,
  title={GLUE: A multi-task benchmark and analysis platform for natural language understanding},
  author={Wang, Alex and Singh, Amanpreet and Michael, Julian and Hill, Felix and Levy, Omer and Bowman, Samuel R},
  journal={arXiv preprint arXiv:1804.07461},
  year={2018}
}

@article{warstadt2019neural,
  title={Neural network acceptability judgments},
  author={Warstadt, Alex and Singh, Amanpreet and Bowman, Samuel R},
  journal={Transactions of the Association for Computational Linguistics},
  volume={7},
  pages={625--641},
  year={2019},
  publisher={MIT Press One Rogers Street, Cambridge, MA 02142-1209, USA journals-info~…}
}

@inproceedings{socher2013recursive,
  title={Recursive deep models for semantic compositionality over a sentiment treebank},
  author={Socher, Richard and Perelygin, Alex and Wu, Jean and Chuang, Jason and Manning, Christopher D and Ng, Andrew Y and Potts, Christopher},
  booktitle={Proceedings of the 2013 conference on empirical methods in natural language processing},
  pages={1631--1642},
  year={2013}
}

@inproceedings{dolan2005automatically,
  title={Automatically constructing a corpus of sentential paraphrases},
  author={Dolan, Bill and Brockett, Chris},
  booktitle={Third international workshop on paraphrasing (IWP2005)},
  year={2005}
}

@article{cer2017semeval,
  title={Semeval-2017 task 1: Semantic textual similarity-multilingual and cross-lingual focused evaluation},
  author={Cer, Daniel and Diab, Mona and Agirre, Eneko and Lopez-Gazpio, Inigo and Specia, Lucia},
  journal={arXiv preprint arXiv:1708.00055},
  year={2017}
}

@article{williams2017broad,
  title={A broad-coverage challenge corpus for sentence understanding through inference},
  author={Williams, Adina and Nangia, Nikita and Bowman, Samuel R},
  journal={arXiv preprint arXiv:1704.05426},
  year={2017}
}

@article{rajpurkar2016squad,
  title={Squad: 100,000+ questions for machine comprehension of text},
  author={Rajpurkar, Pranav and Zhang, Jian and Lopyrev, Konstantin and Liang, Percy},
  journal={arXiv preprint arXiv:1606.05250},
  year={2016}
}

@article{bentivogli2009fifth,
  title={The Fifth PASCAL Recognizing Textual Entailment Challenge.},
  author={Bentivogli, Luisa and Clark, Peter and Dagan, Ido and Giampiccolo, Danilo},
  journal={TAC},
  volume={7},
  number={8},
  pages={1},
  year={2009}
}

@inproceedings{yin20231,
  title={1\% vs 100\%: Parameter-efficient low rank adapter for dense predictions},
  author={Yin, Dongshuo and Yang, Yiran and Wang, Zhechao and Yu, Hongfeng and Wei, Kaiwen and Sun, Xian},
  booktitle={Proceedings of the Computer Vision and Pattern Recognition Conference},
  pages={20116--20126},
  year={2023}
}

@inproceedings{yin20255,
  title={5\%> 100\%: Breaking performance shackles of full fine-tuning on visual recognition tasks},
  author={Yin, Dongshuo and Hu, Leiyi and Li, Bin and Zhang, Youqun and Yang, Xue},
  booktitle={Proceedings of the Computer Vision and Pattern Recognition Conference},
  pages={20071--20081},
  year={2025}
}

@article{fei2006one,
  title={One-shot learning of object categories},
  author={Fei-Fei, Li and Fergus, Robert and Perona, Pietro},
  journal={IEEE Transactions on Pattern Analysis and Machine Intelligence},
  volume={28},
  number={4},
  pages={594--611},
  year={2006},
  publisher={IEEE}
}

@article{krizhevsky2009learning,
  title={Learning multiple layers of features from tiny images},
  author={Krizhevsky, Alex and Hinton, Geoffrey and others},
  year={2009},
  publisher={Toronto, ON, Canada}
}

@inproceedings{cimpoi2014describing,
  title={Describing textures in the wild},
  author={Cimpoi, Mircea and Maji, Subhransu and Kokkinos, Iasonas and Mohamed, Sammy and Vedaldi, Andrea},
  booktitle={Proceedings of the Computer Vision and Pattern Recognition Conference},
  pages={3606--3613},
  year={2014}
}

@inproceedings{nilsback2008automated,
  title={Automated flower classification over a large number of classes},
  author={Nilsback, Maria-Elena and Zisserman, Andrew},
  booktitle={2008 Sixth Indian Conference on Computer Vision, Graphics \& Image Processing},
  pages={722--729},
  year={2008},
  organization={IEEE}
}

@inproceedings{parkhi2012cats,
  title={Cats and dogs},
  author={Parkhi, Omkar M and Vedaldi, Andrea and Zisserman, Andrew and Jawahar, CV},
  booktitle={Proceedings of the Computer Vision and Pattern Recognition Conference},
  pages={3498--3505},
  year={2012},
  organization={IEEE}
}

@inproceedings{xiao2010sun,
  title={Sun database: Large-scale scene recognition from abbey to zoo},
  author={Xiao, Jianxiong and Hays, James and Ehinger, Krista A and Oliva, Aude and Torralba, Antonio},
  booktitle={2010 IEEE Computer Society Conference on Computer Vision and Pattern Recognition},
  pages={3485--3492},
  year={2010},
  organization={IEEE}
}

@inproceedings{netzer2011reading,
  title={Reading digits in natural images with unsupervised feature learning},
  author={Netzer, Yuval and Wang, Tao and Coates, Adam and Bissacco, Alessandro and Wu, Baolin and Ng, Andrew Y and others},
  booktitle={NIPS workshop on deep learning and unsupervised feature learning},
  volume={2011},
  number={5},
  pages={7},
  year={2011},
  organization={Granada}
}

@article{helber2019eurosat,
  title={Eurosat: A novel dataset and deep learning benchmark for land use and land cover classification},
  author={Helber, Patrick and Bischke, Benjamin and Dengel, Andreas and Borth, Damian},
  journal={IEEE Journal of Selected Topics in Applied Earth Observations and Remote Sensing},
  volume={12},
  number={7},
  pages={2217--2226},
  year={2019},
  publisher={IEEE}
}

@article{cheng2017remote,
  title={Remote sensing image scene classification: Benchmark and state of the art},
  author={Cheng, Gong and Han, Junwei and Lu, Xiaoqiang},
  journal={Proceedings of the IEEE},
  volume={105},
  number={10},
  pages={1865--1883},
  year={2017},
  publisher={IEEE}
}

@inproceedings{veeling2018rotation,
  title={Rotation equivariant CNNs for digital pathology},
  author={Veeling, Bastiaan S and Linmans, Jasper and Winkens, Jim and Cohen, Taco and Welling, Max},
  booktitle={International Conference on Medical image computing and computer-assisted intervention},
  pages={210--218},
  year={2018},
  organization={Springer}
}

@article{graham2015kaggle,
  title={Kaggle diabetic retinopathy detection competition report},
  author={Graham, Ben},
  journal={University of Warwick},
  volume={22},
  number={9},
  year={2015}
}

@inproceedings{johnson2017clevr,
  title={Clevr: A diagnostic dataset for compositional language and elementary visual reasoning},
  author={Johnson, Justin and Hariharan, Bharath and Van Der Maaten, Laurens and Fei-Fei, Li and Lawrence Zitnick, C and Girshick, Ross},
  booktitle={Proceedings of the Computer Vision and Pattern Recognition Conference},
  pages={2901--2910},
  year={2017}
}

@inproceedings{lecun2004learning,
  title={Learning methods for generic object recognition with invariance to pose and lighting},
  author={LeCun, Yann and Huang, Fu Jie and Bottou, Leon},
  booktitle={Proceedings of the Computer Vision and Pattern Recognition Conference},
  volume={2},
  pages={II--104},
  year={2004},
}

@article{beattie2016deepmind,
  title={Deepmind lab},
  author={Beattie, Charles and Leibo, Joel Z and Teplyashin, Denis and Ward, Tom and Wainwright, Marcus and K{\"u}ttler, Heinrich and Lefrancq, Andrew and Green, Simon and Vald{\'e}s, V{\'\i}ctor and Sadik, Amir and others},
  journal={arXiv preprint arXiv:1612.03801},
  year={2016}
}

@article{geiger2013vision,
  title={Vision meets robotics: The kitti dataset},
  author={Geiger, Andreas and Lenz, Philip and Stiller, Christoph and Urtasun, Raquel},
  journal={The International Journal of Robotics Research},
  volume={32},
  number={11},
  pages={1231--1237},
  year={2013},
  publisher={Sage Publications Sage UK: London, England}
}

@article{zhang2024neural,
  title={Neural prompt search},
  author={Zhang, Yuanhan and Zhou, Kaiyang and Liu, Ziwei},
  journal={IEEE Transactions on Pattern Analysis and Machine Intelligence},
  volume = {47},
  number = {7},
  pages = {5268--5280},
  year={2024},
  publisher={IEEE}
}

@inproceedings{deng2009imagenet,
  title={Imagenet: A large-scale hierarchical image database},
  author={Deng, Jia and Dong, Wei and Socher, Richard and Li, Li-Jia and Li, Kai and Fei-Fei, Li},
  booktitle={Proceedings of the Computer Vision and Pattern Recognition Conference},
  pages={248--255},
  year={2009},
}

@inproceedings{wu2025unified,
  title={Unified knowledge maintenance pruning and progressive recovery with weight recalling for large vision-language models},
  author={Wu, Zimeng and Chen, Jiaxin and Wang, Yunhong},
  booktitle={Proceedings of the AAAI Conference on Artificial Intelligence},
  volume={39},
  number={8},
  pages={8550--8558},
  year={2025}
}

@inproceedings{zhang2026moismoe,
  title={MP-ISMoE: Mixed-precision interactive side mixture-of-experts for efficient transfer learning},
  author={Zhang, Yutong and Wu, Zimeng and Liao, Shengcai and Wu, Shujiang and Chen, Jiaxin},
  booktitle={Proceedings of the AAAI Conference on Artificial Intelligence},
  volume = {40},
  number = {34},
  pages = {28537--28545},
  year={2026}
}

@article{zhong2026hanwen,
  title={Parameter-Efficient Tuning for Fine-Grained Recognition via Channel-wise Importance Equalization and Diversity Navigation},
  author={Zhong, Hanwen and Chen, Jiaxin and Zhang, Yutong and Huang, Di and Wang, Yunhong},
  journal={IEEE Transactions on Image Processing},
  year={2026},
  publisher={IEEE}
}

@article{chen2026zhenghao,
  title={DAS-SAM: fine-tuning SAM towards drivable area segmentation via efficient multi-scale traffic scene-aware adaptation},
  author={Chen, Zhenghao and Zhou, Nan and Fan, Yi and Zhou, Lina and Xie, Yubao and Chen, Jiaxin and Huang, Di},
  journal={Visual Intelligence},
  volume={4},
  number={1},
  pages={6},
  year={2026},
  publisher={Springer}
}

@inproceedings{zhong2024hanwen,
  title={Transforming vision transformer: Towards efficient multi-task asynchronous learner},
  author={Zhong, Hanwen and Chen, Jiaxin and Zhang, Yutong and Huang, Di and Wang, Yunhong},
  booktitle={Proceedings of the Advances in Neural Information Processing Systems},
  pages={81130--81156},
  year={2024}
}

@inproceedings{zhang2026mingfang,
  title={Parameter-Efficient Adaptation for MLLMs via Implicit Modality Decomposition},
  author={Zhang, Mingfang and Wang, Yunhong and Wang, Lu and Chen, Jiaxin},
  booktitle={Proceedings of the Computer Vision and Pattern Recognition Conference},
  year={2026},
}

@inproceedings{wu2026zimeng,
  title={Collaborative Multi-Mode Pruning for Vision-Language Models},
  author={Wu, Zimeng and Wang, Yunhong and Wang, Donghao and Chen, Jiaxin},
  booktitle={Proceedings of the Computer Vision and Pattern Recognition Conference},
  year={2026},
}
